\documentclass[conference]{IEEEtran}
\usepackage[colorlinks=true,citecolor=blue,linkcolor=blue,urlcolor=blue,bookmarks=true]{hyperref}
\usepackage[noend]{algpseudocode}
\usepackage[numbers, sort]{natbib}
\usepackage{adjustbox}
\usepackage{algorithm}
\usepackage{amsfonts}
\usepackage{amsmath}
\usepackage{amssymb}
\usepackage{array, booktabs, makecell}
\usepackage{babel,blindtext}
\usepackage{braket}
\usepackage{graphicx}
\usepackage{multicol}
\usepackage{multirow}
\usepackage{siunitx}
\usepackage{subcaption}
\usepackage{tabularx}
\usepackage{times}
\usepackage{varwidth}
\usepackage{xspace}
\usepackage{verbatim}
\algrenewcommand\algorithmicindent{1.0em}%

\makeatletter
\newcommand\fs@spaceruled{\def\@fs@cfont{\bfseries}\let\@fs@capt\floatc@ruled
  \def\@fs@pre{\vspace{0.6\baselineskip}\hrule height.8pt depth0pt \kern2pt}%
  \def\@fs@post{\kern2pt\hrule\relax}%
  \def\@fs@mid{\kern2pt\hrule\kern2pt}%
  \let\@fs@iftopcapt\iftrue}
\makeatother

\newcommand{\algacro}{MAGIC\xspace}
\newcommand{\algname}{Macro-Action Generator-Critic\xspace}
\newcommand{\generatornet}{Generator-Net\xspace}
\newcommand{\criticnet}{Critic-Net\xspace}
\newcommand{\planner}{Macro-DESPOT\xspace}

\newcommand{\macroactionset}{\ensuremath{M_\Phi}\xspace}
\newcommand{\macroactionparams}{\ensuremath{\Phi}\xspace}

\newcommand{\ctx}{\ensuremath{c}\xspace}

\newcommand{\generatorparams}{\theta}
\newcommand{\generator}{\ensuremath{G_\generatorparams}\xspace}

\newcommand{\criticparams}{\psi}
\newcommand{\critic}{\ensuremath{\hat{V}_\criticparams}\xspace}
\renewcommand*\Call[2]{\textproc{#1}(#2)}

\newcolumntype{Y}{>{\centering\arraybackslash}X}
\usepackage{xcolor}

\newcommand{\Fbox}[1]{\setlength{\fboxrule}{1.3pt}\setlength{\fboxsep}{0pt}\fbox{#1}}

\usepackage{xcolor}

\newcommand{\secref}[1]{Section~\ref{#1}}
\newcommand{\appendixref}[1]{Appendix~\ref{#1}}
\renewcommand{\eqref}[1]{Eq. (\ref{#1})}
\newcommand{\figref}[1]{Fig.~\ref{#1}}

\renewcommand{\algref}[1]{Algorithm~\ref{#1}}

\newcommand{\tabref}[1]{Table~\ref{#1}}

\newcommand{\ie}{\textit{i.e.}}

\newcommand{\postsubcapspace}{\vspace{-0.3cm}}
\newcommand{\postfigspace}{\vspace{-0.2cm}}

\begin{document}

\title{\LARGE \bf
\algacro: 
Learning Macro-Actions for Online POMDP Planning 
}

\author{Yiyuan Lee, Panpan Cai and David Hsu\\School of Computing, National University of Singapore}

\maketitle

\begin{abstract} 
The partially observable Markov decision process (POMDP) is a principled general framework for robot decision making under uncertainty, but POMDP planning suffers from high computational complexity, when long-term planning is required. 
While temporally-extended \textit{macro-actions} help to cut down the effective planning horizon and significantly improve computational efficiency, how do we acquire good macro-actions? This paper proposes \algname (\algacro), which performs offline learning of macro-actions optimized for online POMDP planning. Specifically, \algacro learns a macro-action \textit{generator} end-to-end, using an online planner's performance as the feedback. 
During online planning, the generator generates on the fly situation-aware macro-actions conditioned on the robot's belief and the environment context. 
We evaluated \algacro on several long-horizon planning tasks both in simulation and on a real robot. The experimental results show that the learned macro-actions offer significant benefits in online planning performance, compared with primitive actions and handcrafted macro-actions.

\end{abstract}

\section{INTRODUCTION}

The partially observable Markov decision process (POMDP) \cite{kaelbling1998pomdp} is a principled approach for planning under uncertainty, and has been successful in numerous real-world robotics settings \cite{morere2017pomdpuavbo,xiao2019pomdpclutter,liu2015pomdpdriving,javdani2018sharedteleop,meghjani2019pomdpdriving}. Many real-world tasks require long-horizon reasoning and involve a continuous action space. However, state-of-the-art online POMDP planners suffer from an exponential complexity w.r.t the planning horizon and the size of the action space. \textit{Macro-actions} serve as a promising tool to cut down the exponential complexity of planning. However, it is challenging to acquire a good, compact set of macro-actions for a given task. Automatically constructing macro-actions is costly, often requiring extensive computation over the state space using carefully-defined criteria \cite{mcgovern2001automatic,mannor2004learningmacro,kurniawati2011migs,ma2015igres,he2010puma}. Moreover, a good choice of macro-actions is often situation-specific, making them hard to handcraft.

We propose to learn situation-aware open-loop macro-actions from data and to plan using these learned macro-actions. This is achieved by learning a macro-action \textit{generator} -- a mapping from the current belief and context (collectively, the online \textit{situation}) to a finite set of macro-actions. The set is then used by a planner to perform efficient online planning. The objective of the learning is to find the optimal macro-action generator that maximizes the downstream planning performance. We argue that this cannot be efficiently solved using typical learning methods: it is hard to manually design effective situation-aware macro-actions for supervised learning; the space of macro-action sets is also exponentially larger than the robot's action space, making exhaustive search or trial-and-error schemes futile.

This paper presents a novel learning method, \textit{\algname} (\algacro) (\figref{magicoverview}), to learn the generator from the feedback of online planning. The key observation is, value estimates output by the planner represent the best task performance achievable when using a macro-action set. By optimizing the generator w.r.t. the value estimates, we learn the generator end-to-end, \textit{directly} for the task performance. 
This is done by learning a differentiable surrogate objective -- a \textit{critic}, to approximate the value function of the planner.
We can then apply the critic to optimize the generator via gradient ascent. After learning offline, the generator is deployed to generate macro-action sets for online planning.

The architecture of \algname can be related to Actor-Critic \cite{konda2000actorcritic}, if we loosely regard the generator as a ``meta-actor'' interacting with both the planner and the physical environment as a joint entity. 
The core difference is, our generator does not learn from raw environment rewards, but from the values fed back by the planner. 

We evaluate \algacro on various long-horizon planning tasks that include continuous action spaces, dynamic and interactive environments, and partial observability. Results show that \algacro learns high-quality macro-actions that lead to superior performance for all tasks, outperforming primitive actions and handcrafted macro-actions. \algacro can efficiently solve large-scale realistic problems such as driving in urban crowds, and has been successfully deployed to a real-world robot platform for mobile manipulation. 

\section{BACKGROUND}

\subsection{POMDP Preliminaries}

A POMDP \cite{kaelbling1998pomdp} is written as a tuple \(\braket{S, A, T, R, \Omega, O, \gamma}\), where \(S, A, \Omega\) denote spaces of states, actions and observations respectively. 
The \textit{reward function} \(R(s,a)\) quantifies the desirability of taking action \(a\) at state \(s\). The \textit{transition function} \(T(s,a,s')=p(s'\mid s,a)\) represents the dynamics of the world, specifying the probability of transiting from \(s\) to \(s'\) by taking action \(a\). The \textit{observation function} \(O(a, s', o)=p(o\mid s',a)\) specifies the probability of observing \(o\) from sensors after taking action \(a\) to reach \(s'\).

POMDPs capture partial observability of the system using a \textit{belief} -- a probability distribution over \(s\), denoted as \(b(s)\). A POMDP policy \(\pi:B\rightarrow A\) prescribes an action for all beliefs in the \textit{belief space} \(B\), the set of all possible probability distributions over \(S\). Solving a POMDP requires finding the \textit{optimal policy} maximizing the expected future reward, or the \textit{value}, for all \(b \in B\).
The value of a policy \(\pi\) at belief \(b\) is defined as:
\begin{equation}
    V_{\pi}(b) = \mathbb{E}\left[\sum_{t = 0}^\infty \gamma^t R(s_t, \pi(b_t)) \mid b_0 = b \right].
\end{equation}

\subsection{Online POMDP Planning}

POMDPs can be solved either \textit{offline} using dynamic programming or \textit{online} using forward search.
Offline POMDP solvers \cite{pineau2003pbvi, smith2005hsvi2, kurniawati2008sarsop, bai2010mcvi} try to generate the full policy for all belief states. However, they suffer heavily from the ``curse of dimensionality'', since the size of the belief space is exponential to that of the state space.

Online POMDP planners \cite{ross2008online,silver2010pomcp} replan at each step, and focus only on reachable contingencies from the current belief. This is typically conducted with a \textit{belief tree search} over a tree, rooted at the current belief and containing all reachable future beliefs within the planning horizon, via recursively branching over actions and observations. State-of-the-art belief tree search planners \cite{silver2010pomcp,kurniawati2016abt,ye2017despot} leverage Monte Carlo sampling and heuristic search to scale up. A representative example is the DESPOT \cite{ye2017despot} algorithm, which samples \(K\) \textit{deterministic scenarios} to approximate the uncertain future. DESPOT reduces the complexity of belief tree search by an exponential factor in the observation space, and can handle continuous states and complex observations. One can also bias action selection using a learned prior policy function \cite{silver2016alphago, silver2017alphagozero, silver2018alphazero, cai2019lets}.

The limitation of DESPOT, however, is the requirement of a small, discretized action space. The complexity of planning with a large or even continuous action space is still considered intractable. 
A possible workaround is to adaptively sample the action space and to progressively widen the belief tree to focus only on useful subsets of the original action space. This class of algorithms includes POMCPOW \cite{sunberg2017pomcpow}, GPS-ABT \cite{seiler2015gpsabt}, QBASE \cite{wang2018qbase}, and CBTS \cite{morere2018cbts}. 
These methods make POMDP planning under continuous action space practical for conceptual or simple real-world problems. 
However, the complexity of planning over a long horizon still prohibits their application to complex real-world tasks.

\subsection{Planning with Macro-Actions}
\textit{Macro-actions} are a form of temporal abstraction to tackle long-horizon planning, by reducing the planning depth linearly and thus the planning complexity exponentially. For deterministic planning and MDPs,
a macro-action can be represented as a sequence of primitive actions or a local trajectory \cite{fikes1972macrooperator,sacerdoti1974planning,korf1983macrooperator,iba1989heuristic}. Such trajectories can be extracted from past experience, typically generated by planning using primitive actions.
A more sophisticated formulation of macro-actions is Semi-MDP (SMDP) \cite{sutton1999smdp}. A macro-action in SMDPs is referred to as an ``option'', a temporally-extended close-loop policy for reaching sub-goals \cite{mcgovern2001automatic,stolle2002learningmacro}, sub-regions \cite{mannor2004learningmacro}, landmarks \cite{mann2015approximate}, or executing low-level skills \cite{konidaris2011autonomous,niekum2013semantically}. A separate high-level planning problem is formulated as an MDP, with an abstract action space consisting of all available options. 

In POMDPs, macro-actions can also be discovered using handcrafted criteria. For offline planning, MIGS \cite{kurniawati2011migs} and IGRES \cite{ma2015igres} pre-build a road-map over the state space by sampling states with high reward gain or information gain, and use the nodes and edges of the road-map to form macro-actions. For online planning, PUMA \cite{he2010puma} uses sub-goal-oriented macro-actions. It first evaluates all states using reward and information gain heuristics. During online belief tree search, it samples the best-evaluated states as sub-goals, and solves the induced local MDPs to produce macro-actions.

Previous algorithms, however, can hardly scale-up to large-scale problems due to the complexity of pre-computing offline macro-actions over the entire state space.
The above algorithms have also decoupled the problem of constructing macro-actions with that of downstream planning. In this paper, we seek to learn open-loop macro-actions optimized \textit{directly} for planning in an end-to-end fashion.

\subsection{Learning to Plan}

The use of learning to improve planning has been a field of recent interest.
A natural and effective approach is to learn planner sub-components, such as dynamics/observation models \cite{hafner2019learning,fang2019dynamics}, heuristics \cite{cai2019letsdrive, pusse2019hyleap}, and sampling distributions \cite{liu2020learningsamplingdistributions}, and to leverage them to scale up planning for complex domains.  
Contrary to prior works, our approach learns action sets for planning -- specifically, a set of open-loop action sequences (or, \textit{macro-actions}) to constrain long-horizon planning under continuous action spaces. 
The idea of learning macro-actions is also related to learning motion primitives or low-level skills \cite{kroemer2016meta, schenck2017learning, wang2021learning, konidaris2018skills}, which can be used for task and motion planning. Here, instead of manually specifying a set of skills and learning them independently, \algacro seeks to automatically discover the most efficient ones, by learning to maximize the planning performance \textit{directly} in an end-to-end fashion.

\section{Overview}

\algacro learns situation-aware macro-actions offline and leverages them for long-horizon online planning. We achieve this by learning a macro-action \textit{generator}  from planning experience. 
\algacro consists of three key components: an online \textit{planner}, a learned macro-action \textit{generator}, and a learned \textit{critic} to facilitate generator learning. 

We represent each macro-action, denoted by \(m\), as an open-loop trajectory, where we assume a suitable parameterization is given according to a task. For instance, 2D Bezier curves (\figref{beziercurve}) can be used for planning with holonomic mobile robots. These trajectories are discretized into sequences of primitive actions for both planning and execution.

Our planner, \planner (\secref{sectionplanning}), conditions online planning on a set of candidate macro-actions, \(\macroactionset\), where \(\macroactionparams\) is the collection of the parameters of all macro-actions in the set. Given a predefined model of the world dynamics and observations, the planner computes an optimal close-loop plan over macro-actions for the robot, depending on the current \textit{belief} \(b\) over world states and the \textit{context} \(c\) of the environment.
At each time step, the planner also outputs a \textit{value estimate} \(v\), which approximates the optimal task performance achievable using online planning when conditioned on the given macro-action set \(\macroactionset\). The key observation is that better macro-action sets will induce higher values. We thus seek to learn macro-actions that maximize this value.

In particular, we learn a \textit{generator}, a neural network, \(\generator\), that outputs a most-suitable macro-action set \(\macroactionset\) for each online situation specified by $(b,c)$ (\secref{sectionlearning}). Such a set is characterized by a uni-modal distribution over \(\macroactionparams\).
We seek to optimize the parameters \(\theta\) of the generator, such that the \textit{expected planner performance},
\begin{equation}\label{eqn:objective}
    \underset{\macroactionparams \sim G_\theta(b, c)}{\mathbb{E}}[V(b, c, \macroactionparams)],
\end{equation}
is maximized at every belief \(b\) and context \(c\), with
the expectation taken over the generator's output distribution. 

The above objective function is, however, very expensive to evaluate for all online situations. We instead learn a \textit{critic}, another neural network \(\critic\), to approximate the planner's value function \(V\).
Using the critic as a differentiable surrogate objective, the generator \(\generator\) is trained to maximize \eqref{eqn:objective} via gradient ascent.

\begin{figure}[t]
    \centering
    \includegraphics[width=\linewidth]{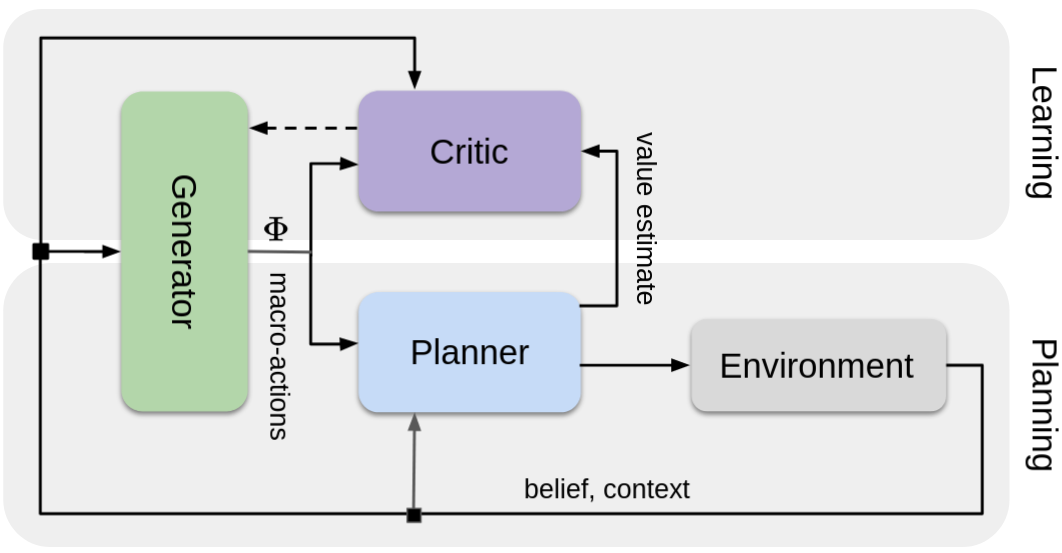}
    \caption{Overview of \algacro. \algacro learns a macro-action \textit{generator}, which prescribes a macro-action set, $\Phi$, for efficient online planning, conditioned on the belief and the environmental context.
    The generator is trained to maximize the \textit{planner}'s performance via gradient ascent (the dashed arrow), using a \textit{critic} as a differentiable surrogate objective.
    }
    \label{magicoverview}
    \postfigspace
\end{figure}

\floatstyle{spaceruled}
\restylefloat{algorithm}
\begin{algorithm}[t]
    \caption{Online planning using learned macro-actions}
    \label{onlinephase}
    \begin{algorithmic}[1]
        \State \(e \leftarrow \Call{InitEnvironment}{\null}\)
        \State \(b \leftarrow \Call{InitBelief}{\null}\)
        \While {\(\Call{IsNotTerminal}{e}\)} 
        \State \(\macroactionparams = \mathbb{E}[\generator(b, c)]\) \label{onlinephase:callgen}
        \State \(m, v \sim \Call{\planner}{b, c, \macroactionparams}\) \label{onlinephase:callplan}
        \For{\(a\) in \(m = (a_1, \dots, a_L)\)}\label{onlinephase:execloop} 
        \State \(e, o \sim \Call{ExecuteAction}{e, a}\) \label{onlinephase:execaction} 
        \State \(b \leftarrow \Call{UpdateBelief}{b, a, o}\) \label{onlinephase:beliefupdate}
        \EndFor
        \EndWhile
    \end{algorithmic}
\end{algorithm}

\figref{magicoverview} shows the workflow of the algorithm in the training phase.
The planner acts episodically, as illustrated in \algref{onlinephase}.
At each time step, the current belief $b$ and context $c$ are first input to the generator to produce parameters $\macroactionparams$ for the candidate macro-action set \(\macroactionset\) (Line \ref{onlinephase:callgen}). The planner then uses \(\macroactionset\) to perform long-horizon planning, taking $b$, \(c\), and \(\Phi\) as inputs, and outputs the optimal macro-action $m \in \macroactionset$ (Line \ref{onlinephase:callplan}). 
The robot executes $m$ (Line \ref{onlinephase:execaction}) and updates the tracked belief using the received observations (Line \ref{onlinephase:beliefupdate}). Additionally, the data point \((b,c,\macroactionparams,v)\) is sent to a \textit{replay buffer}, which holds past experience for learning. 
The learner optimizes both the critic and the generator using gradient ascent. In each iteration, it samples a batch of data from the replay buffer. Using the data, the learner first updates the critic to fit the planner's value estimates; it then updates the generator to maximize for the value approximated by the critic as a differentiable surrogate objective. The updated generator is then used for planning at the next time step. Acting and learning are repeated until convergence or reaching a given limit on training time or data.

At test time, we deploy the generator together with \planner to perform online planning. The process follows \algref{onlinephase}, similarly as in the training phase. 


\section{Online Planning with  Macro-Actions} \label{sectionplanning}

In this section, we present the \planner algorithm in detail, which modifies a state-of-the-art belief tree search algorithm, DESPOT \cite{ye2017despot}, to perform macro-action-based planning. For completeness, we first provide a brief summary of DESPOT, followed by our modifications.

\subsection{DESPOT}
DESPOT samples a set of \textit{scenarios} to represent the current belief and the uncertain future. Each scenario encodes a sampled initial state and a sequence of random numbers that \textit{determinize} the effect of future actions and observations. Using the sampled scenarios, DESPOT constructs a sparse belief tree that ensembles Monte Carlo simulations of the future, generated using a black-box \textit{simulative model} -- a deterministic step function conditioned on the sampled scenarios. Each node in the belief tree contains a set of scenarios whose corresponding states collectively form an approximate representation of a belief. The belief tree starts from the current belief and branches over all actions, but only on observations encountered under the sampled scenarios.
To obtain the optimal policy and its value estimate, DESPOT backs up rewards in the search tree using the Bellman equation:
\begin{eqnarray}\label{eqn:backup}
V(b)=\max_{a \in A}\left\{R(b,a)+\gamma
	\sum_{o\in\Omega}p(o\mid b,a)V(b')\right\}
\end{eqnarray}
where \(b\) and \(b'\) denote an arbitrary belief node and a corresponding child of that node; \(o\) denotes the observation leading to \(b'\); \(V\) is the value estimate maintained for each belief node and \(\gamma\) is the discounting factor that prioritizes short-term returns over long-term returns.

The DESPOT algorithm additionally performs anytime heuristic search to incrementally construct the tree. We refer readers to \cite{ye2017despot} for the precise algorithmic details.

\begin{figure}[!t]
    \centering
    \includegraphics[width=0.8\linewidth]{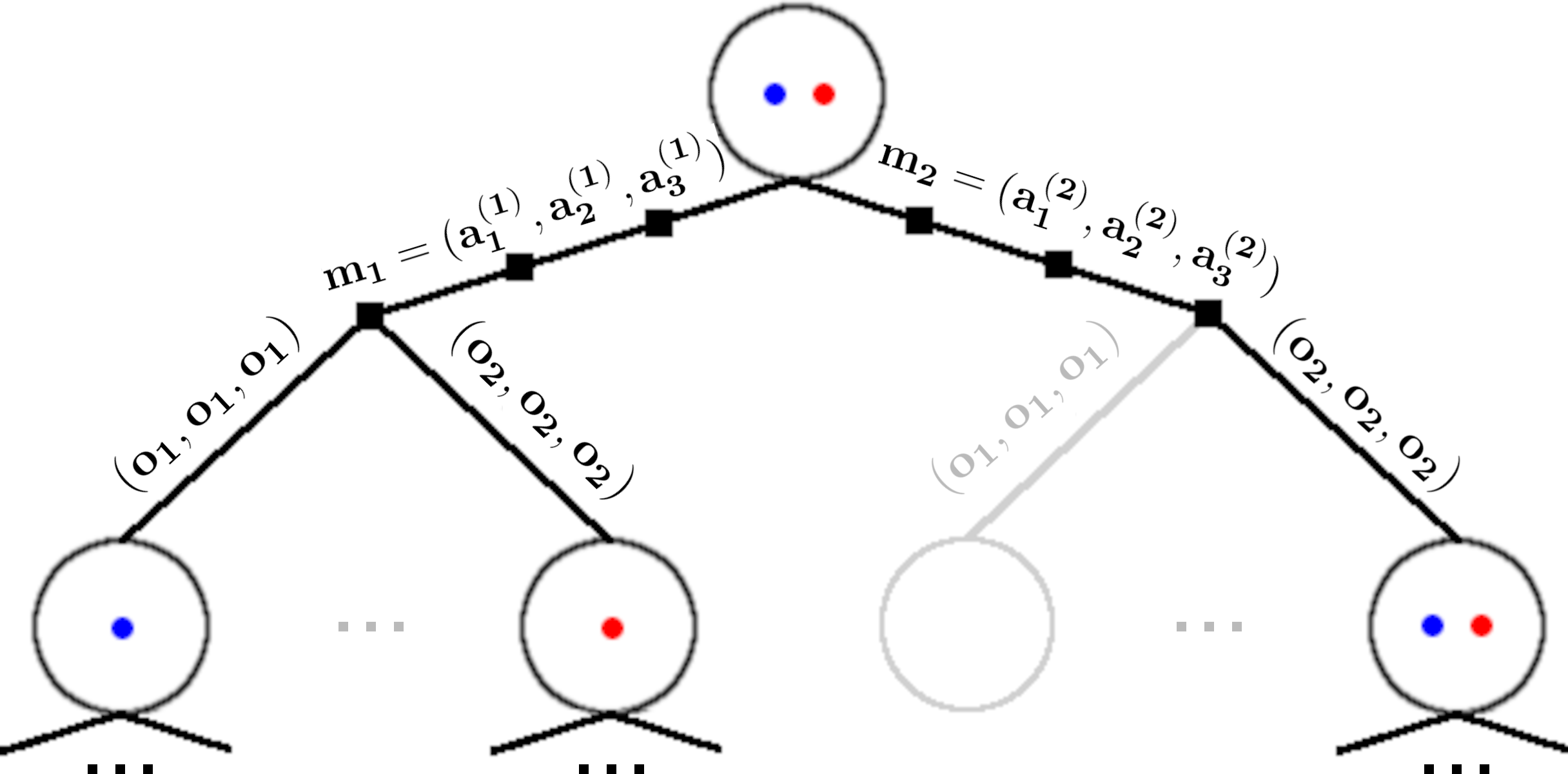}
    \caption{\planner -- online planning using learned macro-actions. A sparse belief tree is searched, branching over possible selections of macro-actions (\(m_1\) and \(m_2\)) and over macro-observations visited under sampled scenarios (red and blue dots). The macro-actions step through multiple actions at once to cut down the exponential complexity.
    }
    \label{figuremacrodespot}
    \postfigspace
\end{figure}

\subsection{Macro-DESPOT}

\planner (\figref{figuremacrodespot}) modifies DESPOT by replacing the primitive action space \(A\) with a set of macro-actions \(\macroactionset\). Under each belief node $b$, the \planner tree branches over all macro-actions \(m\in\macroactionset\). 
Each branch corresponds to a macro-action $m$ being executed entirely in the form of a sequence of primitive actions, \((a_1, \dots, a_L)\), with the environment stepped along the way using the simulative model.
Under each sampled scenario, the macro-action leads to a sequence of observations, \((o_1, \dots, o_L)\), which is treated as a single \textit{macro-observation} \(o \in \Omega^L\). The belief tree further branches with distinct macro-observations encountered under different scenarios, leading to different child belief nodes $b'$.

The backup process in Macro-DESPOT is accordingly modified from the Bellman equation (\eqref{eqn:backup}) to work with macro-actions and macro-observations, yielding
\begin{equation}
\begin{gathered}
    V(b) = \max_{m \in \macroactionset}\left(R(b, m) + \gamma^L \sum_{o\in\Omega^L}  p(o\mid b,m)V(b')\right)
\end{gathered}
\end{equation}
where $R(b, m)=\sum_{i=1}^{L} \gamma^{i - 1} R(b_{i-1}, a_{i})$ represents the cumulative reward along the local trajectories generated by executing $m$, in expectation over $b$;  and where
$p(o\mid b,m) = \prod_{i = 1}^L p(o_i \mid b_{i - 1},a_{i})$ represents the likelihood of receiving the macro-observation $o$ when executing $m$ from $b$. In both $R(b,m)$ and $p(o\mid b,m)$, $b_{i}$ represents the belief updated from $b$ using the partial sequence of actions, $(a_1, \dots, a_{i})$, where $b_0 = b$ is simply the parent belief. 


The belief tree is expanded in an anytime fashion until the allowed planning time is exhausted. Thereafter, \planner outputs the optimal macro-action under the root node to be executed by the robot. During the training phase, it additionally outputs a \textit{value estimate} \(v\) of the optimal macro-action, which approximates the optimal performance that the robot can achieve when constraining online planning to \macroactionset. It can thus be interpreted as a \textit{quality measure} of \macroactionset for the given situation, and is used as the learning signal to train the macro-action generator. 


\begin{figure*}[t]
    {\centering
    \setlength{\fboxsep}{0pt}
    \renewcommand{\tabcolsep}{2pt}

    \begin{tabularx}{\linewidth}{ccc}
        \Fbox{\includegraphics[height=107.5px]{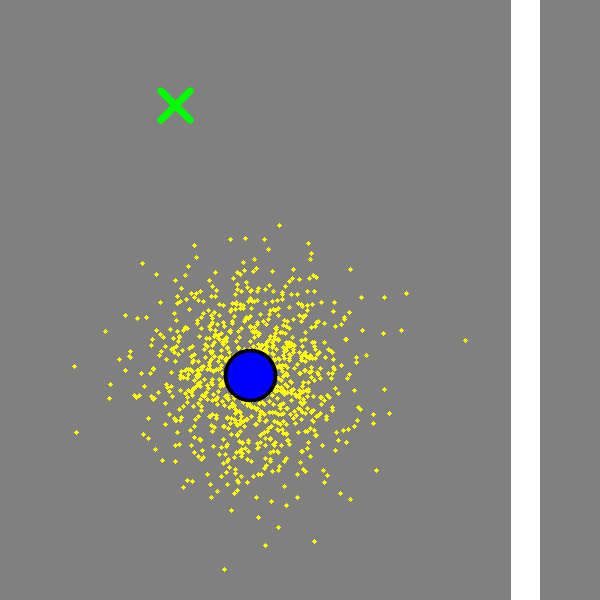}}
        &
        \Fbox{\includegraphics[height=107.5px]{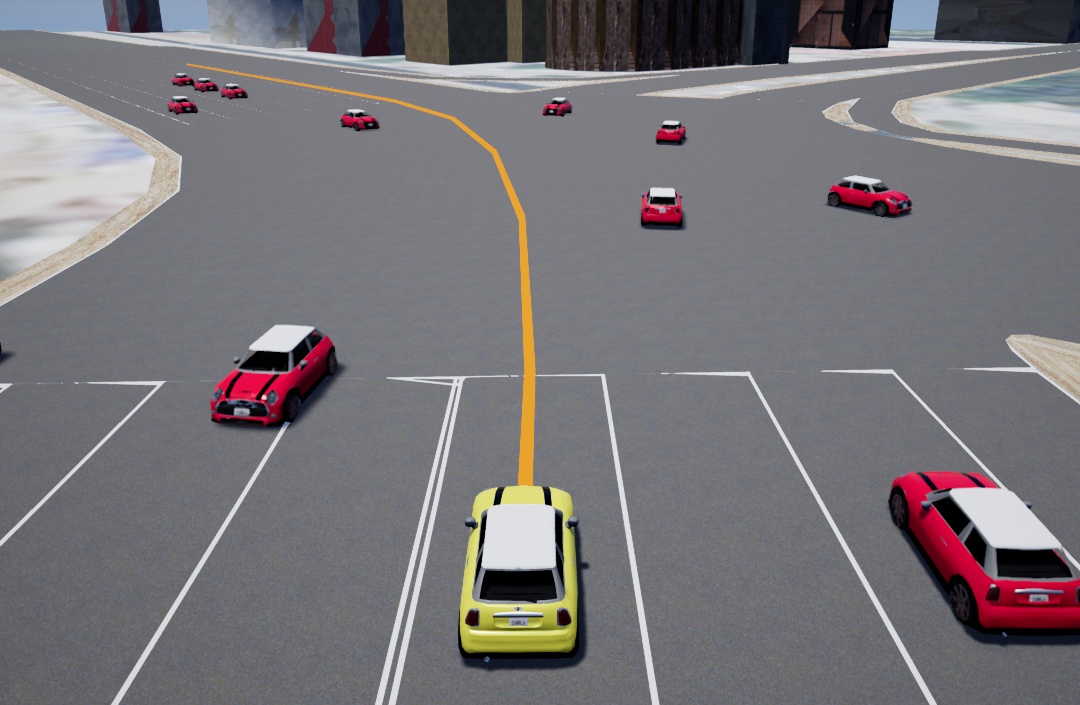}}
        &
        \Fbox{\includegraphics[height=107.5px]{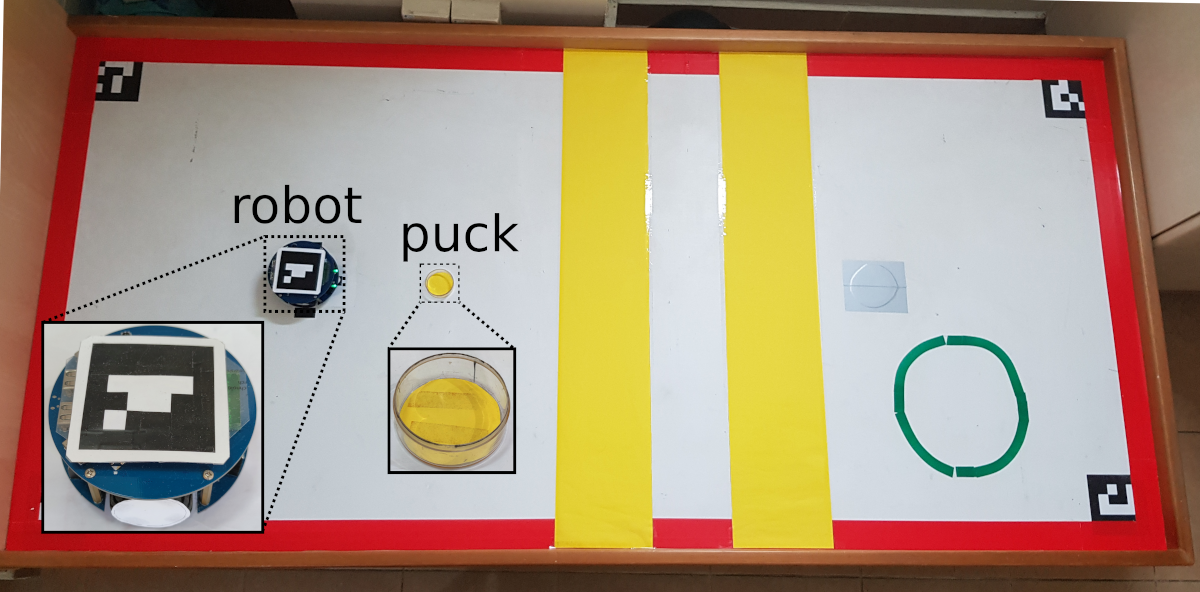}}
        \\
        \begin{subfigure}{80px}\subcaption{Light-Dark}\label{macrodemolightdark}\end{subfigure}&
        \begin{subfigure}{80px}\subcaption{Crowd-Driving}\label{macrodemocrowddrive}\end{subfigure}&
        \begin{subfigure}{80px}\subcaption{Puck-Push}\label{macrodemopuckpush}\end{subfigure}
    \end{tabularx}
    }
    \postsubcapspace
    \caption{\small (a) Light-Dark task. The robot (blue) navigates to a goal (green) through darkness (gray background), using the light (white strip) to localize itself. Particles show the robot's belief over its position. \small(b) Crowd-Driving task. The robot vehicle (yellow) drives through a busy intersection, making maneuvers to follow the intended path (orange) steadily without colliding with the other vehicles (red). \small (c) Puck-Push task. The robot pushes a plastic puck (yellow) to a goal region (dark green circle), past two visually-occluding regions (yellow strips).}
    \label{macrodemo}
    \postfigspace
\end{figure*}

\begin{figure*}[t]
    {\centering
    \setlength{\fboxsep}{0pt}
    \renewcommand{\tabcolsep}{2pt}
    \begin{tabularx}{\linewidth}{YYY}
        \multicolumn{3}{c}{\includegraphics[width=0.6\linewidth]{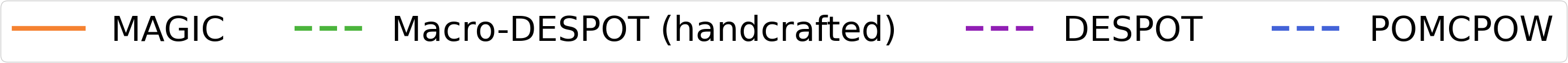}}\\
        \includegraphics[width=\linewidth]{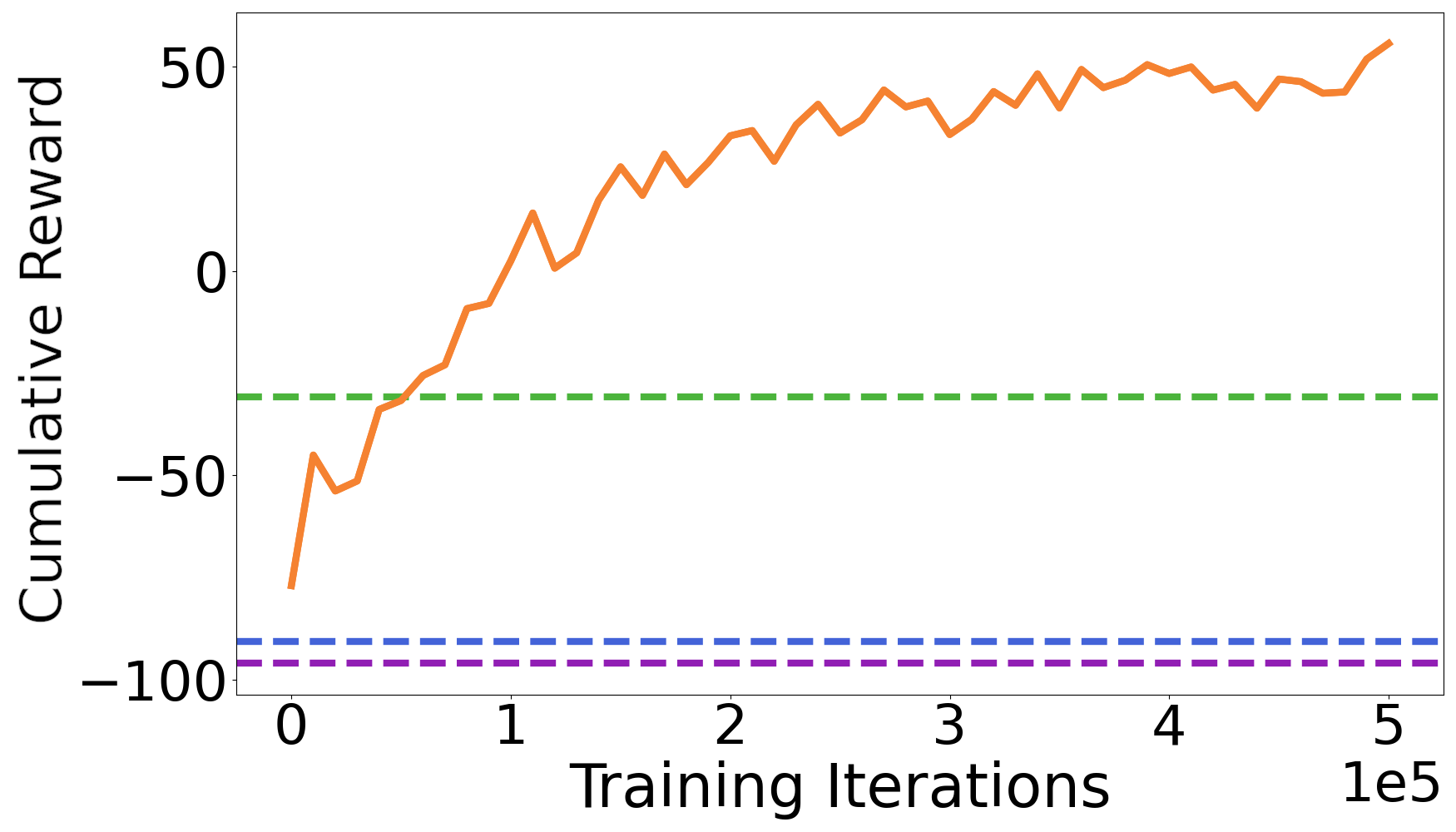}&
        \includegraphics[width=\linewidth]{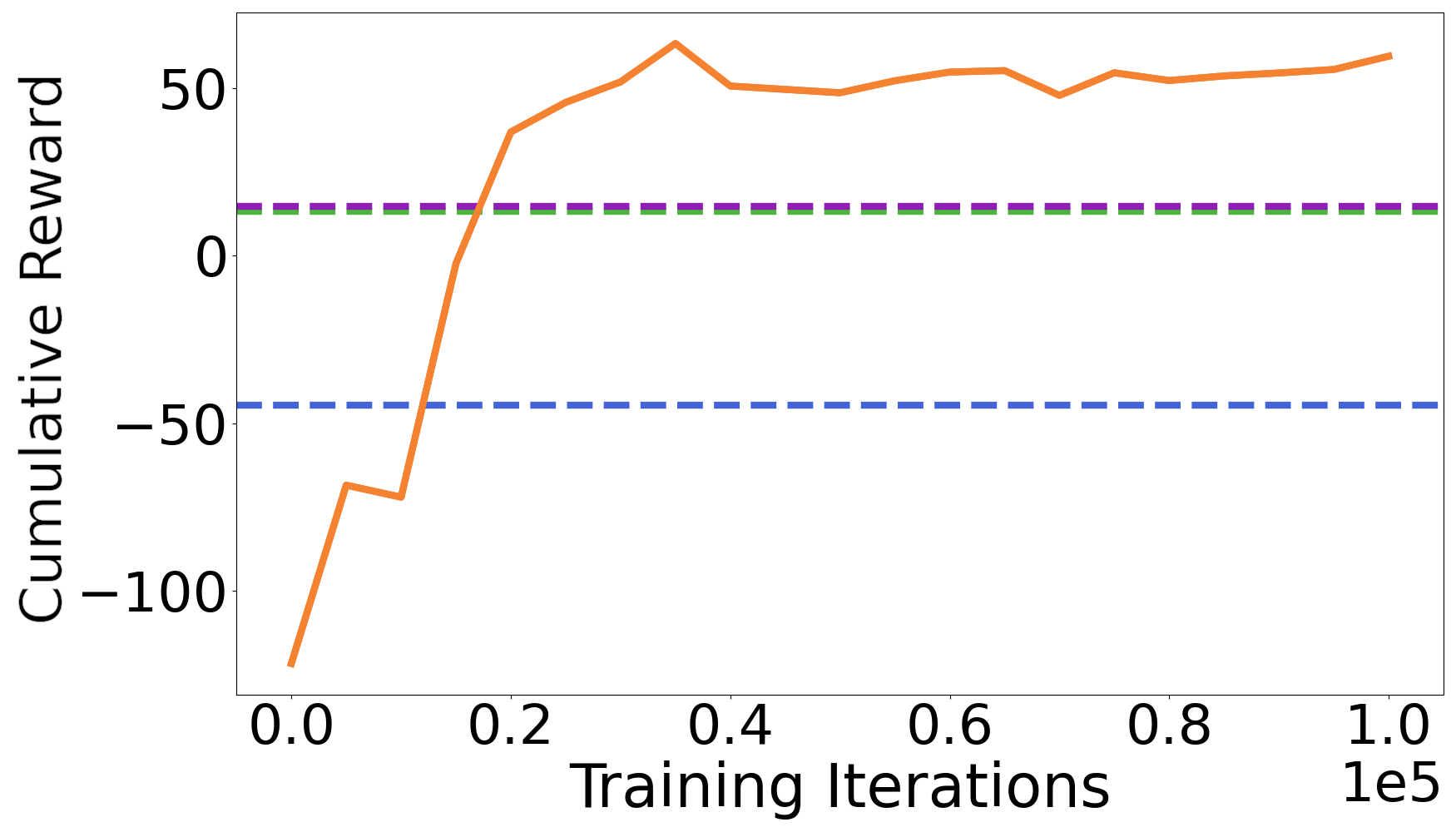}&
        \includegraphics[width=\linewidth]{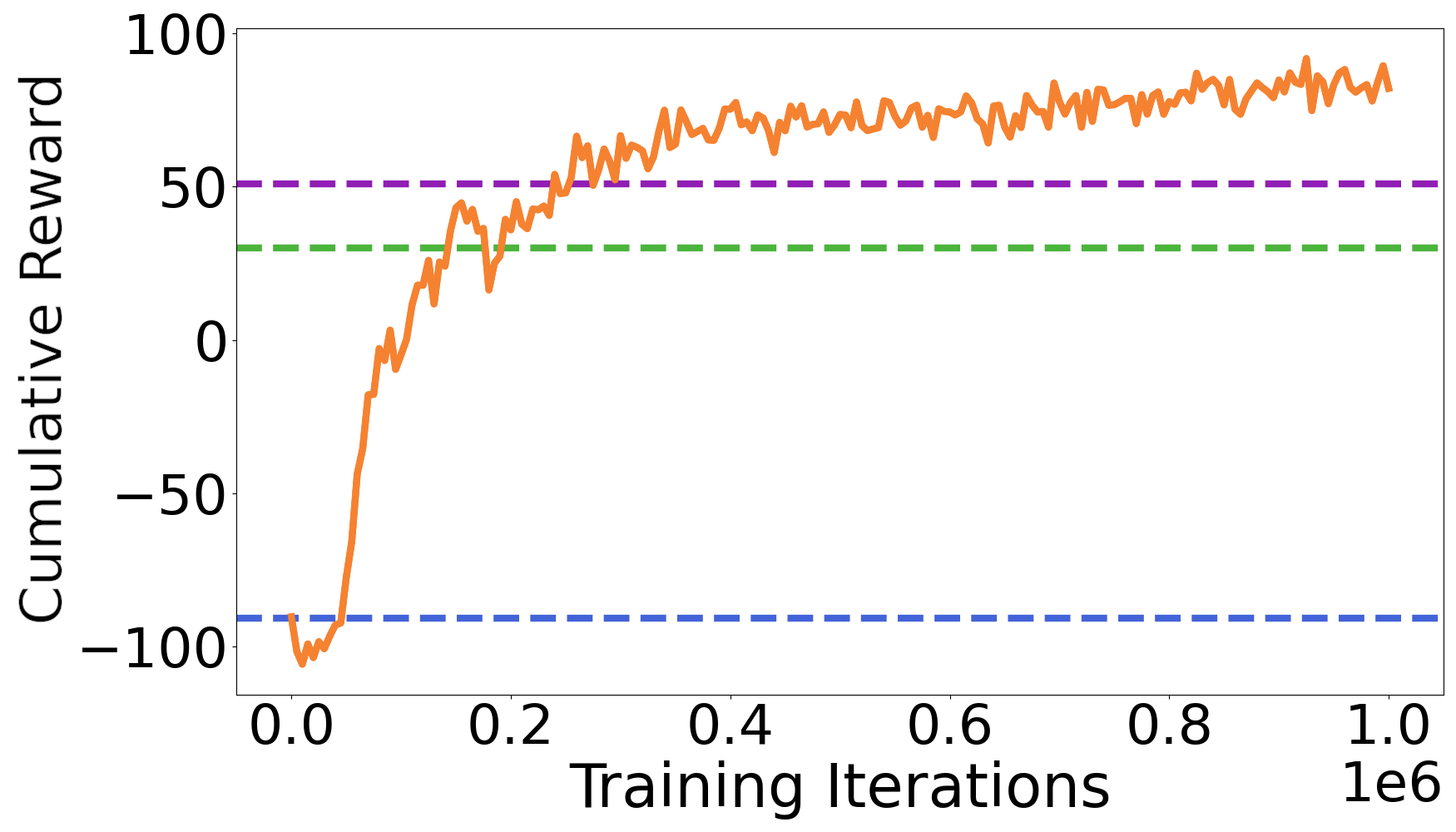}
        \\
        \begin{subfigure}{80px}\subcaption{Light-Dark}\label{fig:curve_lightdark}\end{subfigure}&
        \begin{subfigure}{80px}\subcaption{Crowd-Driving}\label{fig:curve_crowddrive}\end{subfigure}&
        \begin{subfigure}{80px}\subcaption{Puck-Push}\label{fig:curve_puckpush}\end{subfigure}
    \end{tabularx}
    }
    \postsubcapspace
    \caption{Performance (cumulative reward) of MAGIC over training progress for {\small (a)} Light-Dark, {\small (b)} Crowd-Driving, and {\small (c)} Puck-Push. Horizontal lines show the performances of the baselines for comparison.}
    \postfigspace
\end{figure*}

\section{Learning Macro-Actions for Online Planning} \label{sectionlearning}

We propose a novel method to learn an optimal generator that maximizes the expected planning performance defined in \eqref{eqn:objective}. We decompose the learning problem into two stages. First, we learn a \textit{critic} function, \(\critic\), to model the core element in \eqref{eqn:objective} -- the value function of the planner, \(V\). Then, we train the \textit{generator}, \(\generator\), using the critic as a surrogate objective, to approximately maximize \eqref{eqn:objective}. The critic and the generator are trained jointly from scratch, and will converge toward the true objective (\eqref{eqn:objective}) and its optimal solution, respectively, when continuously provided with data under proper optimization.

We represent both the critic and the generator as neural networks, and train them using the planner's experience via gradient ascent. 
We represent an experience or a data point as \((b, \ctx, \macroactionparams, v)\), where \(b\) is the input belief, \(\ctx\) is the context, \(\macroactionparams\) denotes parameters of the macro-action set used for planning, and \(v\) is the optimal value as estimated by \planner. 

Due to the stochastic nature of the planner's value estimates, we use a Gaussian distribution to represent each \(\critic(b, c, \macroactionparams)\). We also make the generator's output \(\generator(b, \ctx)\) distributional, parameterized as a Gaussian, to enable random exploration at training time.
Network architectures used for the critic and the generator are presented in \appendixref{appendixnnarchitectures} and shown in \figref{figurenncriticnet} and \figref{figurenngeneratornet}, respectively.

\subsection{Learning the Critic}
\algacro directly fits the critic to the planner's value function, by minimizing, for all \(b\), \(\ctx\), and \(\macroactionparams\), the KL-divergence between the output distribution of the critic, $\critic(b, \ctx, \macroactionparams)$, and the actual distribution of the planner's value estimates. 
Particularly, given a data set \(D\), \algacro updates \critic to maximize the log-likelihood of the observed value estimates:
\begin{equation} \label{eqncriticobjective}
    J(\psi) = \underset{(b, \ctx, \macroactionparams, v) \sim D}{\mathbb{E}} [\log p_{\psi}(v)],
\end{equation}
where $p_{\psi}$ denotes the probability density function corresponding to \(\critic(b, \ctx, \macroactionparams)\). 

\subsection{Learning the Generator}
\algacro optimizes the generator to output, for any $(b,c)$, a distribution over macro-action set parameters $\macroactionparams$, that maximizes the expected planning performance as estimated by the critic. Given a data set $D$, we update \generator using gradient ascent, to maximize the following objective:
\begin{equation} \label{eqnobjective}
     \begin{gathered}
     J(\theta) = \underset{(b, \ctx) \sim D}{\mathbb{E}}\left[ \underset{\macroactionparams \sim \generator(b, \ctx)}{\mathbb{E}}\left[
     \mathbb{E}[\critic(b, \ctx, \macroactionparams)]\right] 
      + 
      \alpha 
      \mathcal{H}(\generator(b, \ctx))\right].
    \end{gathered}
\end{equation}

The first term is a surrogate objective approximating \eqref{eqn:objective} and allows gradients to propagate back to the generator. 
We include an additional entropy regularization term over the generator's output distribution, \(\generator(b, \ctx)\), which is used to enforce a minimum level of exploration during training. When the regularization weight $\alpha$ is gradually annealed to zero, the above objective will converge to Eqn. (\ref{eqn:objective}).


The original form of \eqref{eqnobjective} contains a sampling process which is not differentiable. To fix this, we reparameterize \(\generator\) as a deterministic function \(f_{\theta}(\epsilon; b, \ctx)\) conditioned on a random seed \(\epsilon \sim \mathcal{N}(0,1)\). The gradient  of \eqref{eqnobjective} is thus rewritten as:
\begin{equation} \label{eqngeneratorgrad}
\begin{aligned}
    &&\nabla_{\theta} J(\theta) = \underset{(b, \ctx) \sim D}{\mathbb{E}} \left[
    \underset{\epsilon \sim \mathcal{N}(0, 1)}{\mathbb{E}}\left[
    \nabla_{\theta} J(\theta)|_{(b,c,\epsilon)}
    \right]\right]
\end{aligned}
\end{equation}
where $\nabla_{\theta} J(\theta)|_{(b,c,\epsilon)}$ is the point-wise gradient calculated as:
\begin{equation} \label{eqnchainrule}
\begin{aligned}
    \nabla_{\macroactionparams} \mathbb{E}[\critic(b, \ctx, \macroactionparams))]\nabla_{\generatorparams}f_{\generatorparams}(\epsilon; b, \ctx) - \alpha\cdot  \nabla_{\generatorparams} \log p_{\theta}(\macroactionparams)
\end{aligned}
\end{equation}
with \(\macroactionparams = f_{\theta}(\epsilon; b, \ctx)\), and $p_{\theta}$ being the probability density function of the distribution \(\generator(b, \ctx)\) output by the generator.
The first term of \eqref{eqnchainrule} applies the chain rule to calculate the gradient of \eqref{eqnobjective} w.r.t. the generator's parameters, $\theta$.

The regularization weight \(\alpha\) in \eqref{eqnobjective} is automatically adjusted using gradient ascent to fulfill a desired minimum target entropy \(\mathcal{H}_0\) for the generator's output distributions. This is done by maximizing \(\alpha\) with the following objective:
\begin{equation} \label{alphaupdate}
    J(\alpha) = \alpha\cdot\underset{(b, \ctx) \sim D}{\mathbb{E}} \left[\mathcal{H}_0 -  \mathcal{H}(\generator(b^{(i)}, \ctx^{(i)}))\right].
\end{equation}
Intuitively, \(\alpha\) is increased when the expected entropy of the generator's output falls below the desired target $\mathcal{H}_0$, and decreased otherwise. $\alpha$ is kept non-negative with a rectifier.

Application of the chain rule and entropy regularization is inspired by a model-free RL algorithm, SAC \cite{haarnoja2018sac}, where an entropy-regularized soft-Q function is learned as a surrogate objective to help optimize a policy. Here, our ``policy'' instead generates an entire macro-action set for downstream planning, and the learning signals are not raw rewards from the environment, but the value estimates from a planner. 

\begin{figure*}[t]
    {\centering
    \renewcommand{\tabcolsep}{5pt}
    \begin{tabularx}{\linewidth}{YYYYY}
        \Fbox{\includegraphics[width=\linewidth]{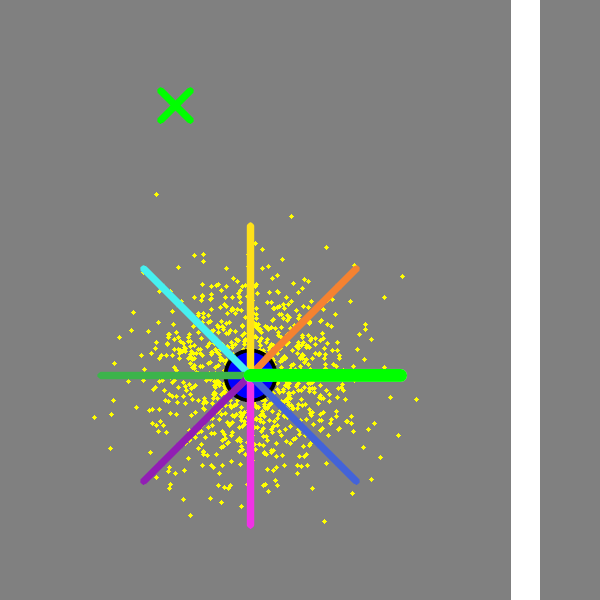}}&
        \Fbox{\includegraphics[width=\linewidth]{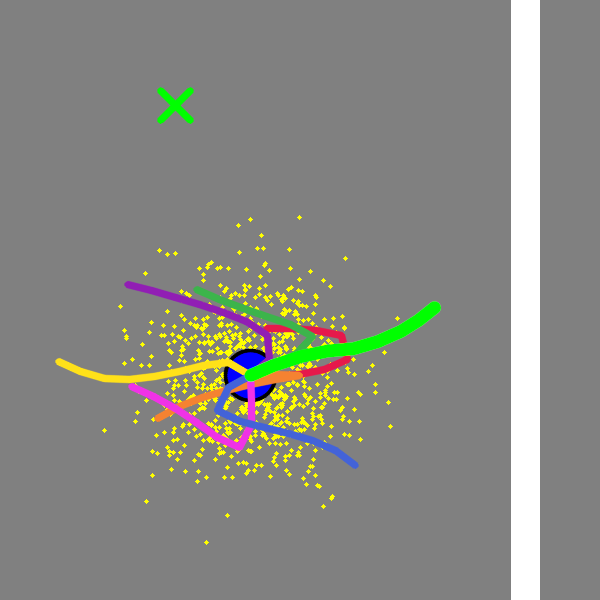}}&
        \Fbox{\includegraphics[width=\linewidth]{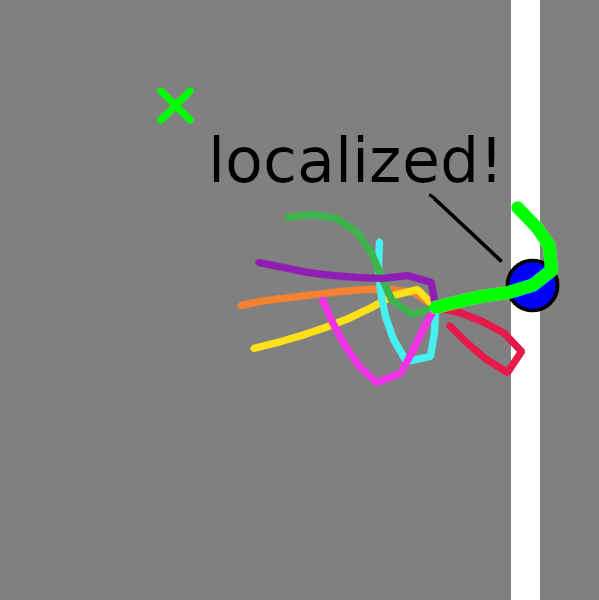}}&
        \Fbox{\includegraphics[width=\linewidth]{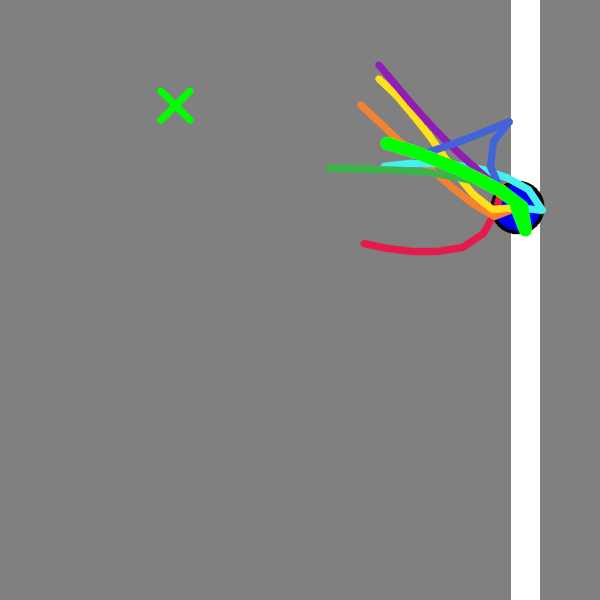}}&
        \Fbox{\includegraphics[width=\linewidth]{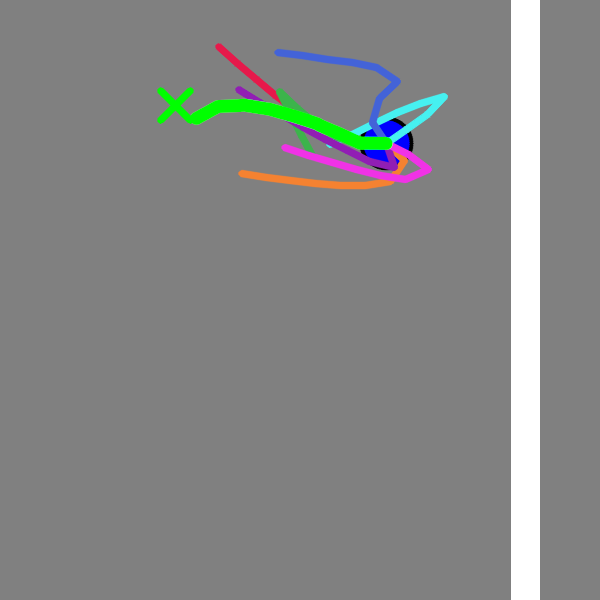}}
        \\
        \begin{subfigure}{80px}\subcaption{\null}\label{lightdarkdemoa}\end{subfigure}&
        \begin{subfigure}{80px}\subcaption{\null}\label{lightdarkdemob}\end{subfigure}&
        \begin{subfigure}{80px}\subcaption{\null}\label{lightdarkdemoc}\end{subfigure}&
        \begin{subfigure}{80px}\subcaption{\null}\label{lightdarkdemod}\end{subfigure}&
        \begin{subfigure}{80px}\subcaption{\null}\label{lightdarkdemoe}\end{subfigure}
    \end{tabularx}
    }
    \postsubcapspace
    \caption{Handcrafted (a) and learned (b-e) macro-actions for Light-Dark, with the belief over the robot's position shown as yellow particles and the optimal macro-actions chosen by the planner highlighted in bright green: (b) episode starts; (c) robot localized; (d-e) learned macro-actions concentrate toward the goal post-localization.}
    \label{lightdarkdetailed}
    \postfigspace
\end{figure*}

\section{Experiments} \label{sec::experiments}

In our experiments, we aim to answer a few questions:
\begin{enumerate}
    \item How much do the learned macro-actions benefit online POMDP planning?
    \item Can \algacro generalize to novel situations and transfer to realistic environments?
    \item Can \algacro scale up to complex tasks?
    \item What has \algacro learned in its macro-action set?
\end{enumerate}

We apply \algacro to a set of virtual and real-world tasks, including Light-Dark (\figref{macrodemolightdark}), a classical POMDP benchmark requiring active information gathering; Crowd-Driving (\figref{macrodemocrowddrive}), a realistic task involving interactions with many other agents; and Puck-Push (\figref{macrodemopuckpush}), a real-world robot mobile manipulation task. These tasks cover important challenges in robot planning such as partial observability, acting and sensing uncertainties, complex interactions, and large-scale environments. Each task has a continuous action space and critically requires long-horizon reasoning. 

To understand the advantage of planning using macro-actions, we first compare \algacro with primitive-action-based planners. These include \textit{DESPOT} \cite{ye2017despot}, which uses discretized primitive actions to perform belief tree search; and \textit{POMCPOW} \cite{sunberg2017pomcpow}, which progressively samples primitive actions from the continuous action space. 
To further understand the advantage of \textit{learning} macro-actions, we also compare \algacro with \textit{\planner (handcrafted)}, which uses handcrafted macro-actions instead of learned ones.

\begin{table}[t]
    {\centering
    \caption{Performance (cumulative reward) of \algacro and planning algorithms across all evaluation tasks. Numbers in brackets represent standard errors (throughout this paper).}
    \begin{tabularx}{\linewidth}{cYYY}
    \toprule
     & Light-Dark & \begin{tabular}[c]{@{}c@{}}Crowd-Driving\\ (POMDP Sim.)\end{tabular} & \begin{tabular}[c]{@{}c@{}}Puck-Push\\ (POMDP Sim.)\end{tabular} \\ \midrule

    \begin{tabular}[c]{@{}c@{}}MAGIC\end{tabular} & 
    \textbf{54.1 (1.0)} & 
    \textbf{58.6 (2.0)} & 
    \textbf{87.9 (1.0)} \\

    \begin{tabular}[c]{@{}c@{}}Macro-DESPOT\\ (Handcrafted)\end{tabular} & 
    -30.9 (1.0) & 
    13.1 (2.0) & 
    34.0 (2.0) \\

    DESPOT & 
    -96.1 (1.0) & 
    14.8 (2.0) & 
    53.0 (2.0) \\

    POMCPOW & 
    -90.7 (0.5) & 
    -44.7 (1.0) & 
    -94.1 (1.0) \\ 
    
    \bottomrule

    \end{tabularx}
    \label{tablefinal}
    \postfigspace
    }
\end{table}

Our results show that macro-actions tremendously benefit long-horizon planning over continuous action spaces. By efficiently learning the macro-actions, \algacro fully exploits this benefit, achieving the best planning performance for all evaluation tasks. 
\algacro successfully scales up on the Crowd-Driving task, which has an
enormous state space and complex dynamics. \algacro also generalizes well to novel environments. While \algacro is trained completely in randomized simulations built upon POMDP models, it successfully transfers to a realistic simulator for Crowd-Driving and a real-world setup for Puck-Push.

By visualizing the learned macro-actions, we show that \algacro learns to cover both reward-exploiting and information-gathering behaviors, which help to focus the search tree. \algacro also learns to adapt the macro-action sets to different beliefs and contexts, further improving the performance of online planning. 

Additional visual results are demonstrated in this video: \url{https://leeyiyuan.info/links/magic-rss21}.

\subsection{Training and Planning Performance}\label{subsectionperformance}
\figref{fig:curve_lightdark}-\subref{fig:curve_puckpush} show the learning curves of \algacro on the three evaluation tasks. As the generator is randomly initialized, the initial performance is relatively low. During training, \algacro quickly learns to generate useful macro-actions and begins to outperform the best of the other approaches. This happens at around \(1.5 \times 10^5, 2\times10^4\), and \(2.5 \times 10^5\) updates for Light-Dark, Crowd-Driving, and Puck-Push respectively, corresponding to around \(2\), \(0.4\), and \(3.5\) hours of training using a single Nvidia RTX 2080 GPU with batch-size 256 and 16 workers. 

\begin{table}[t]
    \centering
    \caption{Detailed performance of \algacro and planning algorithms on Light-Dark.}
    \begin{tabularx}{\linewidth}{ccccc}
    \toprule
  
        & \begin{tabular}[c]{@{}c@{}}Acc.\\ Reward\end{tabular} & \begin{tabular}[c]{@{}c@{}}Success\\ Rate (\%)\end{tabular} & \begin{tabular}[c]{@{}c@{}}Steps \\Taken\footnotemark \end{tabular} & \begin{tabular}[c]{@{}c@{}}Min. Tracking\\ Error\end{tabular} \\ \midrule

        MAGIC & 
        \textbf{54.1 (1.0)} & 
        \textbf{79.0 (0.5)} & 
        37.6 (0.1) & 
        \textbf{0.28 (0.01)} \\

        \begin{tabular}[c]{@{}c@{}}Macro-DESPOT\\ (handcrafted)\end{tabular} & 
        -30.9 (1.0) &
        37.1 (0.5) &
        44.7 (0.1) &
        0.67 (0.01) \\

        DESPOT & 
        -96.1 (1.0) & 
        4.9 (0.5) & 
        59.8 (0.2) & 
        1.59 (0.01) \\

        POMCPOW & 
        -90.7 (0.5) & 
        6.2 (0.3) & 
        \textbf{22.1 (0.4)} & 
        1.55 (0.01) \\
    
    \bottomrule
    \end{tabularx}
    \label{tablelightdark}
    \postfigspace
    \end{table}
\footnotetext{Calculated over successful runs.}

\begin{figure*}[!t]
    {\centering
    \setlength{\fboxsep}{0pt}
    \renewcommand{\tabcolsep}{3pt}
    \begin{tabularx}{\linewidth}{YYYY}
        \fbox{\includegraphics[width=\linewidth]{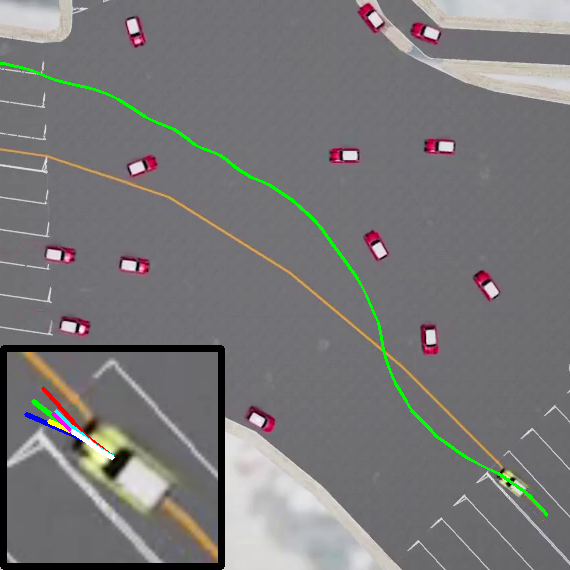}}&
        \fbox{\includegraphics[width=\linewidth]{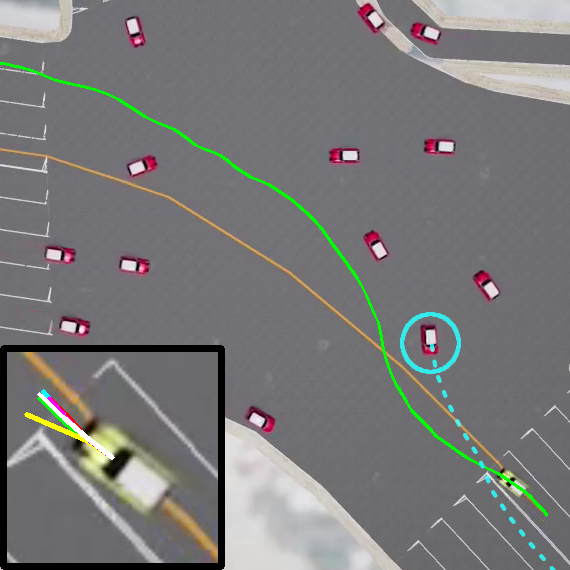}}&
        \fbox{\includegraphics[width=\linewidth]{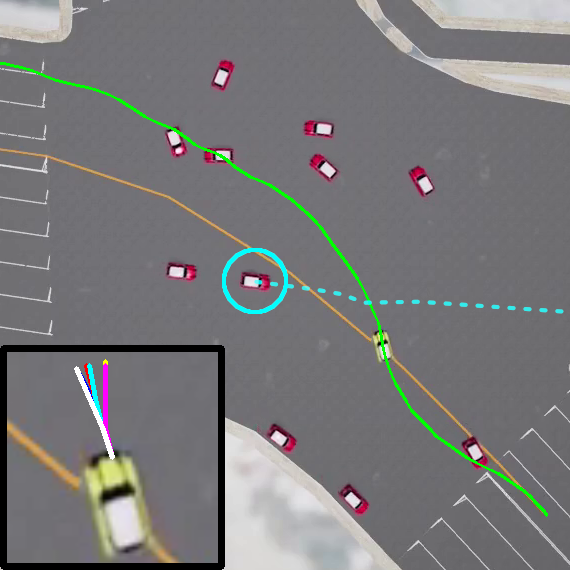}}&
        \fbox{\includegraphics[width=\linewidth]{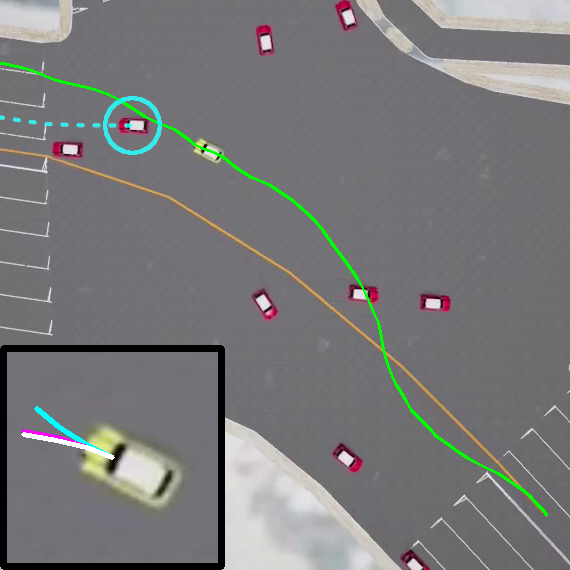}}\\
        \begin{subfigure}{80px}\subcaption{\null}\label{crowddrivedemoa}\end{subfigure}&
        \begin{subfigure}{80px}\subcaption{\null}\label{crowddrivedemob}\end{subfigure}&
        \begin{subfigure}{80px}\subcaption{\null}\label{crowddrivedemoc}\end{subfigure}&
        \begin{subfigure}{80px}\subcaption{\null}\label{crowddrivedemod}\end{subfigure}
    \end{tabularx}
    }
    \postsubcapspace
    \caption{Handcrafted (a) and learned (b-d) macro-actions for Crowd-Driving. The target path is shown in orange. The final trajectory of the robot is shown in green: (b) avoiding incoming vehicle; (c) cutting past incoming vehicle; (d) overtaking.
    }
    \label{crowddrivedemo}
    \postfigspace
\end{figure*}

\begin{table*}[!t]
    \centering
    \caption{Detailed performance of \algacro and planning algorithms on Crowd-Driving in both the POMDP simulator and the SUMMIT simulator.}
 \begin{tabular}{ccccccccc}
\toprule
 &
  \multicolumn{4}{c}{POMDP Simulator} &
  \multicolumn{4}{c}{SUMMIT} \\
  \cmidrule(lr){2-5}\cmidrule(lr){6-9}
 &
  \begin{tabular}[c]{@{}c@{}}Acc. \\ Reward\end{tabular} &
  \begin{tabular}[c]{@{}c@{}}Distance\\ Covered\end{tabular} &
  \begin{tabular}[c]{@{}c@{}}Stall\\ Rate\end{tabular} &
  \begin{tabular}[c]{@{}c@{}}Search \\ Depth\end{tabular} &
  \begin{tabular}[c]{@{}c@{}}Acc. \\ Reward\end{tabular} &
  \begin{tabular}[c]{@{}c@{}}Distance \\ Covered\end{tabular} &
  \begin{tabular}[c]{@{}c@{}}Stall\\ Rate\end{tabular} &
  \begin{tabular}[c]{@{}c@{}}Search \\ Depth\end{tabular} \\ \midrule
MAGIC &
  \textbf{58.6 (2.0)} &
  \textbf{111.0 (1.0)} &
  \textbf{0.069 (0.001)} &
  \textbf{22.5 (0.1)} &
  \textbf{32.2 (1.0)} &
  97.2 (1.0) &
  \textbf{0.100 (0.001)} &
  \textbf{18.3 (0.2)} \\
Macro-DESPOT (handcrafted) &
  13.1 (2.0) &
  98.1 (1.0) &
  0.099 (0.001) &
  18.0 (0.2) &
  4.9 (1.0) &
  \textbf{101.5 (1.0)} &
  0.285 (0.001) &
  \textbf{18.0 (0.2)} \\
DESPOT &
  14.8 (2.0) &
  91.8 (1.0) &
  \textbf{0.071 (0.001)} &
  17.6 (0.1) &
  -8.2 (2.0) &
  66.7 (1.0) &
  0.259 (0.002) &
  11.8 (0.5) \\
POMCPOW &
  -44.7 (1.0) &
  70.4 (1.0) &
  0.130 (0.001) &
  2.0 (0.1) &
  -18.3 (1.0) &
  75.3 (1.0) &
  0.312 (0.001) &
  1.9 (0.1) \\ \bottomrule
\end{tabular}
    \label{tablecrowddrive}
    \postfigspace
\end{table*}  

\tabref{tablefinal} shows the overall planning performance of \algacro, in terms of the average cumulative reward of episodes, as compared to standard planning algorithms. Planning with macro-actions (\algacro and \planner (handcrafted)) generally outperforms planning with primitive ones (DESPOT and POMCPOW), because all the evaluation tasks critically require long-horizon policies. Among the macro-action-based planners, \algacro significantly outperforms handcrafted macro-actions. This is achieved by learning macro-actions which directly maximize the planning performance. The improvements are prominent on all tasks. For Light-Dark, the learned macro-actions help to perform long-term information gathering and precise goal seeking; For Crowd-Driving, they help to tackle the large-scale dynamic environment; For Puck-Push, they assist the robot to reliably manipulate the puck using complex interactions. 

Note that \algacro relies entirely on the POMDP model for both planning and learning, and did not require additional domain knowledge compared to learn good macro-actions. Conceptually, the training process is analogous to the hyperparameter tuning of online planners, where, in contrast, our hyperparameters are the macro-action sets specific to the online situations, which cannot be efficiently tuned in standard ways. \algacro offers a principled and efficient way to do so.

Next, we investigate the benefits of \algacro on each evaluation task independently.

\subsection{Light-Dark -- active information gathering} \label{subsectionexplightdark}

\subsubsection{Task}
\textit{Light-Dark} (\figref{macrodemolightdark}) is adapted from the classical POMDP benchmark in \cite{platt2010lightdark}. The robot (blue) attempts to move to and stop \textit{precisely} at a goal (green) in the dark area, where the robot's position can only be observed within the light area (white), thus requiring active information gathering. The goal position, light position, and initial distribution of the robot are \textit{randomly generated} for each episode. The former two constitute the context \(\ctx\) of the environment. See a detailed description of the task in \appendixref{appendixtaskdetails}.

The optimal policy for Light-Dark is to localize within the light before moving to the goal. The light is constrained to be far from the robot's initial position, requiring long-horizon reasoning to discover the value of such active information-gathering actions.

\begin{figure*}[!t]
    {\centering
    \setlength{\fboxsep}{0pt}
    \renewcommand{\tabcolsep}{3pt}
    \begin{tabularx}{\linewidth}{YYYY}
        \fbox{\includegraphics[width=\linewidth]{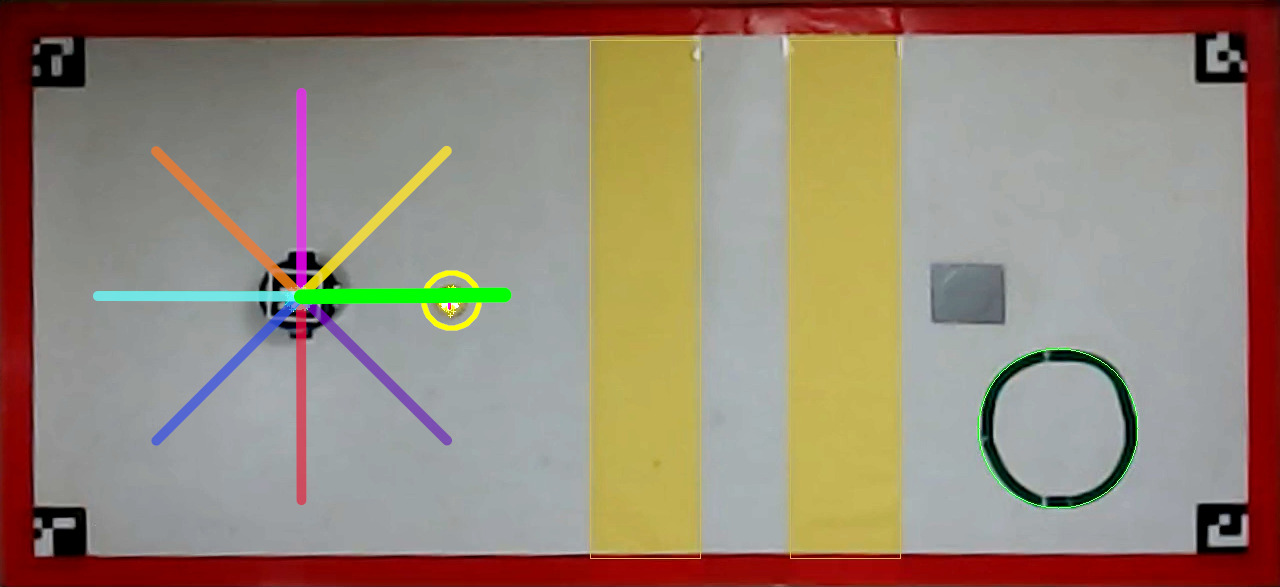}}&
        \fbox{\includegraphics[width=\linewidth]{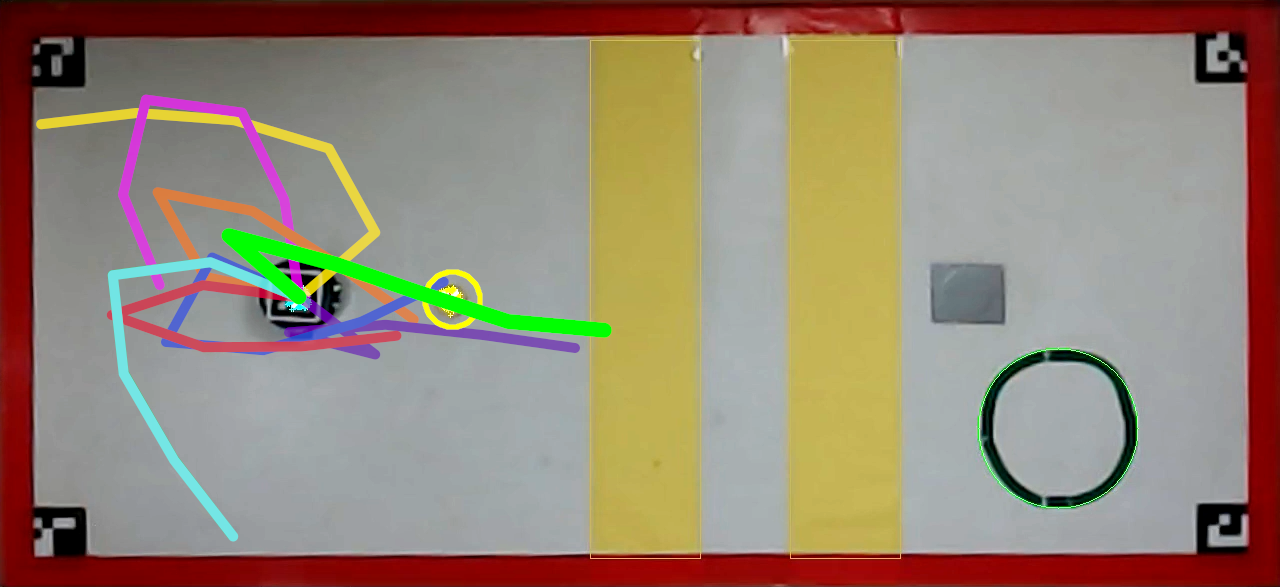}}&
        \fbox{\includegraphics[width=\linewidth]{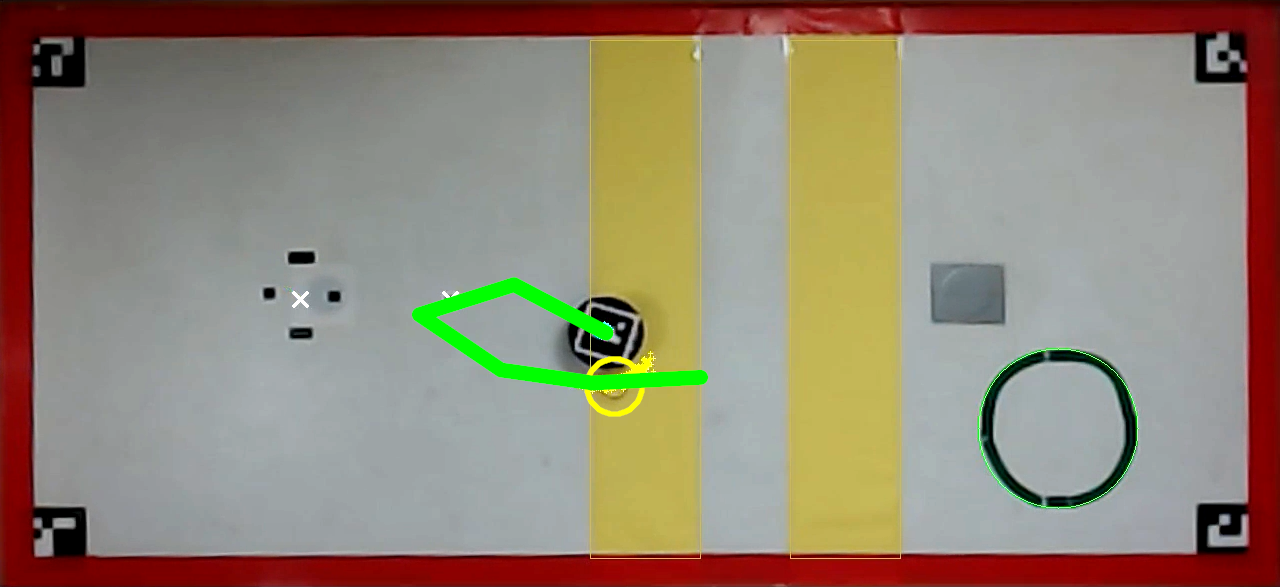}}&
        \fbox{\includegraphics[width=\linewidth]{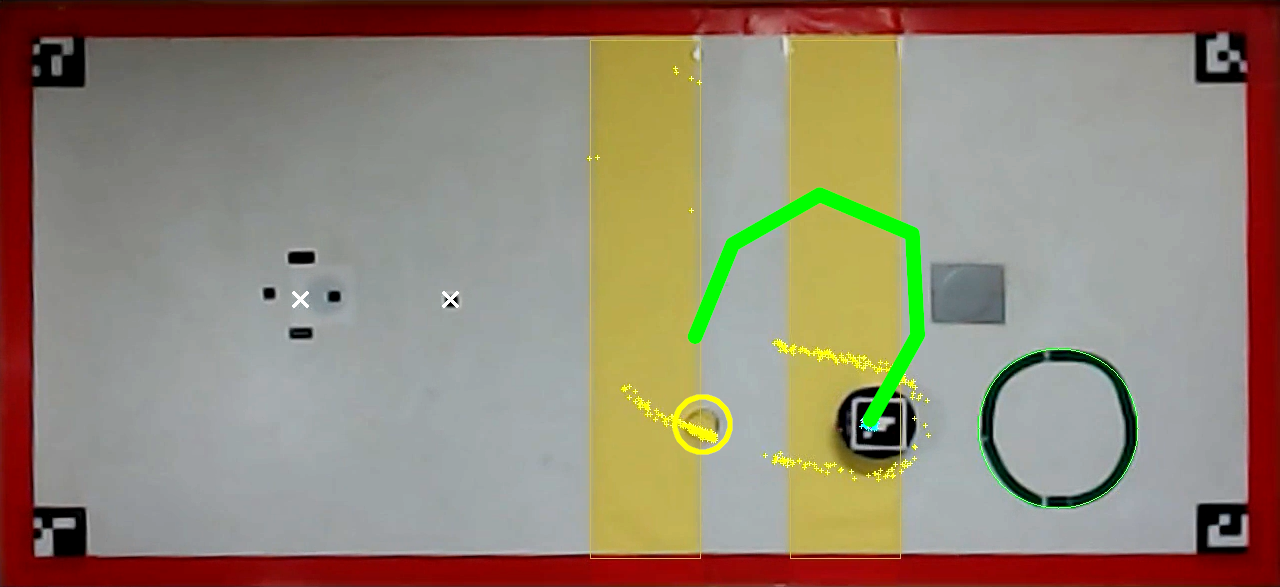}}\\
        \begin{subfigure}{80px}\subcaption{\null}\label{puckpushdemoa}\end{subfigure}&
        \begin{subfigure}{80px}\subcaption{\null}\label{puckpushdemob}\end{subfigure}&
        \begin{subfigure}{80px}\subcaption{\null}\label{puckpushdemoc}\end{subfigure}&
        \begin{subfigure}{80px}\subcaption{\null}\label{puckpushdemod}\end{subfigure}
    \end{tabularx}
    }
    \postsubcapspace
    \caption{Handcrafted (a) and learned (b-d) macro-actions for Puck-Push, with the belief shown as yellow particles and the optimal macro-actions chosen by the planner highlighted in bright green: (b) episode starts; (c) reversing to re-push; (d) exploring the occlusion region to re-localize the puck.}
    \label{puckpushdemo}
    \postfigspace
\end{figure*}

\begin{table*}[!t]
    \centering
    \caption{Detailed performance of \algacro and planning algorithms on Puck-Push evaluated on both the POMDP simulator and the real-world setup.}
    \begin{tabular}{ccccccccc}
    \toprule
    \multicolumn{1}{c}{} & \multicolumn{4}{c}{POMDP Simulator} & \multicolumn{4}{c}{Real-World Setup} \\
    \cmidrule(lr){2-5}\cmidrule(lr){6-9}
    \multicolumn{1}{c}{} & \begin{tabular}[c]{@{}c@{}}Acc. \\ Reward\end{tabular} & \begin{tabular}[c]{@{}c@{}}Success \\ Rate (\%)\end{tabular} & \begin{tabular}[c]{@{}c@{}}Steps \\ Taken\footnotemark[1]\end{tabular} & \multicolumn{1}{c}{\begin{tabular}[c]{@{}c@{}}Search\\ Depth\end{tabular}} & \begin{tabular}[c]{@{}c@{}}Acc. \\ Reward\end{tabular} & \begin{tabular}[c]{@{}c@{}}Success \\ Rate (\%)\end{tabular} & \begin{tabular}[c]{@{}c@{}}Steps \\ Taken\footnotemark[1]\end{tabular} & \begin{tabular}[c]{@{}c@{}}Search\\ Depth\end{tabular} \\ \midrule

MAGIC &
  \textbf{87.9 (1.0)} &
  \textbf{95.3 (0.5)} &
  35.3 (0.1) &
  \textbf{89.4 (0.1)} &
  \textbf{90.9 (5.2)} &
  \textbf{97.5 (2.5)} &
  39.4 (2.0) &
  \textbf{83.1 (2.5)} \\
Macro-DESPOT (handcrafted) &
  34.0 (2.0) &
  70.0 (1.0) &
  42.1 (0.6) &
  61.0 (0.4) &
  9.6 (15.9) &
  58.5 (7.8) &
  57.1 (4.6) &
  51.3 (4.2) \\
DESPOT &
  53.0 (2.0) &
  78.6 (1.0) &
  \textbf{26.2 (0.6)} &
  69.1 (0.5) &
  3.5 (16.4) &
  55.0 (8.0) &
  35.9 (6.2) &
  64.1 (4.5) \\
POMCPOW &
  \multicolumn{1}{l}{-94.1 (1.0)} &
  7.6 (0.5) &
  31.0 (1.7) &
  \multicolumn{1}{c}{2.1 (0.1)} &
  -84.1 (11.0) &
  12.5 (5.3) &
  27.8 (13.9) &
  4.2 (3.0) \\
    
    \bottomrule
    \end{tabular}
    \label{tablepuckpush}
    \postfigspace
\end{table*}

\subsubsection{Macro-action parameterization}
We use 2D cubic Bezier curves (\figref{beziercurve}) to represent macro-actions in Light-Dark. Each macro-action is parameterized by \(6\) real numbers, corresponding to the \(3\) control points of the curve. A macro-action is executed by scaling and discretizing the curve into fixed-length line segments, each corresponding to a primitive action; the primitive actions are then executed consecutively. Each macro-action set contains \(8\) macro-actions each of length \(8\). \(\macroactionparams\) is thus a vector of \(6 \times 8 = 48\) real numbers, obtained by concatenating the control points of all curves.
The handcrafted macro-actions (\figref{lightdarkdemoa}) are straight-line trajectories of length \(6\) pointing to \(8\) general directions. All macro-action lengths are selected using hyperparameter search.

\subsubsection{Analysis}
\tabref{tablelightdark} shows the detailed performance of the tested algorithms. \algacro achieves the highest success rate, leading to the best cumulative rewards. A crucial observation is that \algacro maintains more concentrated beliefs over the robot's position than other algorithms, as shown by the smaller minimum tracking error achieved within episodes.
As visualized in \figref{lightdarkdetailed}, the above advantages are achieved by learning to cover information-gathering trajectories pointing to the light in the starting phase, and to concentrate on goal-seeking trajectories that exploit reward after the robot is well-localized. 
In contrast, by planning with uniformly-directed handcrafted macro-actions, the robot struggles to reach the goal precisely; primitive-action-based planners such as DESPOT and POMCPOW frequently fail to gather information and do not localize using the light, due to insufficient search depths.

\subsection{Crowd-Driving -- complex environments}\label{subsectionexpcrowddrive}
\subsubsection{Task}
In \textit{Crowd-Driving} (\figref{macrodemocrowddrive}), a robot vehicle drives along a target path at a busy, unregulated intersection through crowded traffic, seeking to achieve both safety and efficiency when controlled using speed and steering commands. 
The robot can observe the positions, orientations, and velocities of all vehicles, but requires inferring the exo-agents' intended paths, or \textit{intentions}, from interaction history. 
All vehicles are \textit{randomly spawned} over the map, seeking to cross the intersection. The context \(\ctx\) of the environment is the ego-vehicle's target path, which is also randomly selected. A detailed description of the task is available in \appendixref{appendixtaskdetails}.

The Crowd-Driving task has an enormous state space, consisting of the joint state of all nearby exo-vehicles, inducing a complex planning problem. Additionally, the uncertain intentions, non-holonomic kinematics, and high-order dynamics of traffic participants require the planner to perform sophisticated long-horizon reasoning to avoid collisions and to progress efficiently. 

\subsubsection{Macro-action parameterization}
We use \textit{turn-and-go} curves (\figref{turnandgocurve}) to represent macro-actions in Crowd-Driving. Each macro-action is parameterized by \(2\) real numbers, \(speed\) and \(steer\). The first half of the curve turns with a fixed steering angle, \(steer\), and a constant command speed, \(speed\); the second half drives straight, using the same command speed and with no steering. Actual trajectories of the vehicle are subject to its kinematics and dynamics.
Each macro-action set contains \(7\) macro-actions each of length \(3\). \(\macroactionparams\) is thus a vector of \(2 \times 7 = 14\) real numbers, obtained by concatenating the parameters of the individual macro-actions.
The handcrafted macro-actions (\figref{crowddrivedemoa}) are parameterized in the same way, but uniformly covering trajectories with maximum and zero steering, as well as maximum, half-of-maximum, and zero command speeds. 

\subsubsection{Analysis}
We train \algacro using a world simulator built with the above POMDP model, and test \algacro on a high-fidelity simulator, SUMMIT \cite{cai2020summit} (\figref{macrodemocrowddrive}), that features realistic physics. Both simulators use a real-world map replicating the Meskel Square intersection in Addis Ababa, Ethiopia. 

\tabref{tablecrowddrive} shows the performance of \algacro on both the POMDP simulator and on SUMMIT. \algacro offers the best driving policy among all algorithms, safely traveling the longest distances with the least amount of stalling. 
This combination of safety and efficiency can only be achieved by reacting to potential collisions ahead of time. We observe that \algacro generally searches deeper than other algorithms, as the learned macro-actions induce more focused search.

\algacro also successfully transfers to random crowds on the SUMMIT simulator, despite the non-negligible differences in the environment dynamics. Consistent with the POMDP results, \algacro significantly outperforms planning using primitive actions and handcrafted macro actions, thus achieving safer and more efficient driving.

\figref{crowddrivedemo} provides examples of the learned macro-actions. \algacro learns to include in its macro-action set three types of local trajectories -- those that deviate from the target path to avoid exo-vehicles; those that recover back to the path; and those that drive straight with a high speed whenever possible to maximize the driving smoothness. \algacro adapts the set to different online situations. For instance, in \figref{crowddrivedemob} and \ref{crowddrivedemoc}, the sets are biased toward smooth driving and collision avoidance with incoming vehicles, while in \figref{crowddrivedemod}, it concentrates on ways to overtake the slow-moving vehicle in front and to subsequently return to the target path.

\subsection{Puck-Push -- real robot experiment}\label{subsectionexppuckpush}

In \textit{Puck-Push} (\figref{macrodemopuckpush}), the robot attempts to \textit{push} a puck from a start position to a goal region. 
The robot acts similarly as in Light-Dark, except that it pushes the puck around upon contact. The puck has a relatively smooth surface, and thus slides across the robot's circumference when being pushed. 
The robot can observe the position of itself and the puck with slight noise via a bird-view camera, except when the puck passes two \textit{occluding regions} (yellow), where the puck's color fuses into the background and thus becomes undetectable. 
The robot and the puck's initial positions are fixed, but the goal is \textit{randomly generated}. 
The context \(\ctx\) is thus the goal position. See a detailed description of the task in \appendixref{appendixtaskdetails}.

Puck-Push requires fine interactions with the puck, and mistakes have long-term consequences: if the puck slides past the robot completely during some particular push, the robot is required to navigate back to re-push it toward the goal; if the puck reaches the boundary, it would be impossible to be pushed to the goal. Long-horizon planning is thus critical.

\subsubsection{Macro-action parameterization}
We use 2D Bezier curves (\figref{beziercurve}), similar to Light-Dark, to parameterize macro-actions in Puck-Push. Each macro-action set contains \(8\) macro-actions each of length \(5\). \(\macroactionparams\) is thus a vector of \(6 \times 8 = 48\) real numbers, obtained by concatenating the control points of all curves.
The handcrafted macro-actions (\figref{puckpushdemoa}) are straight-line trajectories of length \(2\) pointing to \(8\) general directions. All macro-action lengths are selected using hyperparameter search.

\subsubsection{Analysis}

We train \algacro solely on the POMDP simulator and test it on a real-world setup (\figref{macrodemopuckpush}). Unlike in simulation, the goal position in the real-world experiment is fixed. Physical coefficients in the POMDP model are manually calibrated using real-world data.

\tabref{tablepuckpush} shows the performance of \algacro on both the POMDP simulator and on the real robot with comparisons to other planning algorithms. \algacro achieves a remarkably higher success rate. Compared to planning with primitive actions, \algacro can avoid unrecoverable mistakes, such as pushing the puck toward the wall, through effective long-horizon reasoning. Compared to the handcrafted macro-actions (\figref{puckpushdemoa}), \algacro learns much more flexible local trajectories (\figref{puckpushdemob}-\subref{puckpushdemod}), allowing reliable interactions with the puck.
Particularly, the learned macro-actions include different ways to manipulate the puck, either by pushing it directly toward the goal (\figref{puckpushdemob}), or by retracting to re-push it whenever the puck slides past the robot (\figref{puckpushdemoc}), or by exploring the occluding regions to re-localize the puck (\figref{puckpushdemod}). Such diverse and meaningful macro-actions help the planner to discover flexible strategies and search significantly deeper, thus achieving superior performance.

Furthermore, \algacro successfully transfers to the real-world setup with unstable camera observations and real-world dynamics, and can push the puck to the goal in almost all trials. This transfer capability is enabled by simulating the stochasticity of the real-world using properly randomized transitions and observations in the POMDP model, and applying the randomization in both planning and learning.

\begin{table}[t]
    {\centering
    \caption{Ablation study of MAGIC by using only certain situational information.}
    \begin{tabularx}{\linewidth}{cYYY}
    \toprule
    & Light-Dark & \begin{tabular}[c]{@{}c@{}}Crowd-Driving\\ (POMDP Sim.)\end{tabular} & \begin{tabular}[c]{@{}c@{}}Puck-Push\\ (POMDP Sim.)\end{tabular} \\ \midrule
    Belief + Context & 
    \textbf{54.1 (1.0)} & 
    \textbf{58.6 (2.0)} & 
    \textbf{87.9 (1.0)} \\

    Belief Only & 
    -51.2 (1.0) & 
    \textbf{53.7 (2.0)} & 
    82.5 (1.0) \\

    Context Only & 
    35.1 (1.0) & 
    \textbf{56.6 (2.0)} &
    66.2 (1.0) \\

    Unconditioned & 
    43.9 (1.0) & 
    15.9 (2.0) & 
    64.3 (1.0) \\

    \bottomrule
    \end{tabularx}
    \label{tableablation}
    \postfigspace
    }
\end{table}

\begin{figure}[t]
    {\centering
    \setlength{\fboxsep}{0pt}
    \renewcommand{\tabcolsep}{3pt}
    \begin{tabularx}{\linewidth}{YYY}
        \Fbox{\includegraphics[width=\linewidth]{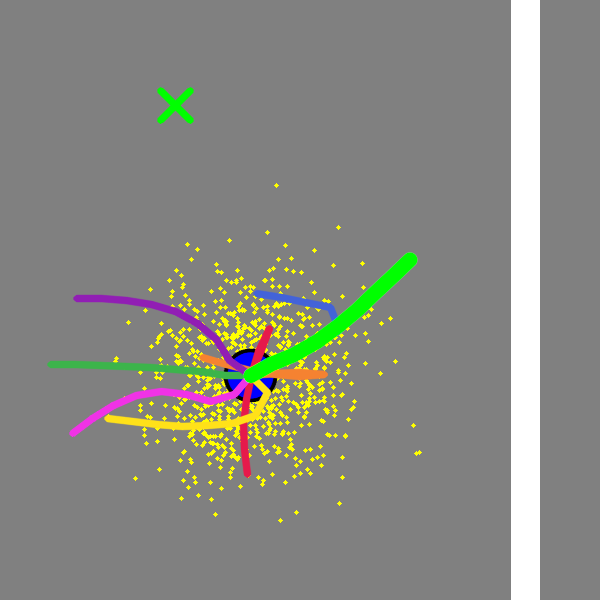}}&
        \Fbox{\includegraphics[width=\linewidth]{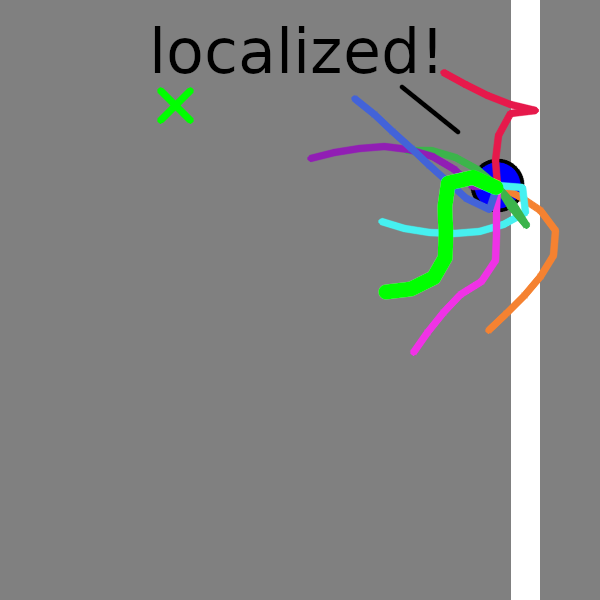}}&
        \Fbox{\includegraphics[width=\linewidth]{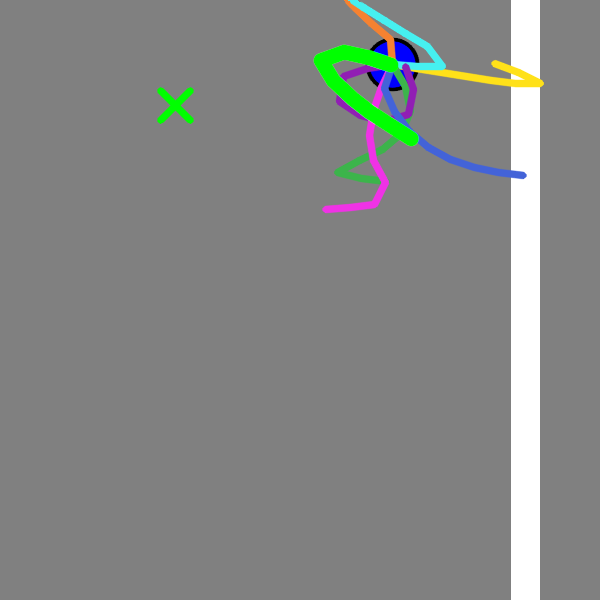}}
        \\
        \begin{subfigure}{80px}\subcaption{\null}\label{lightdarkoverfita}\end{subfigure}&
        \begin{subfigure}{80px}\subcaption{\null}\label{lightdarkoverfitb}\end{subfigure}&
        \begin{subfigure}{80px}\subcaption{\null}\label{lightdarkoverfitc}\end{subfigure}
        \\
        \Fbox{\includegraphics[width=\linewidth]{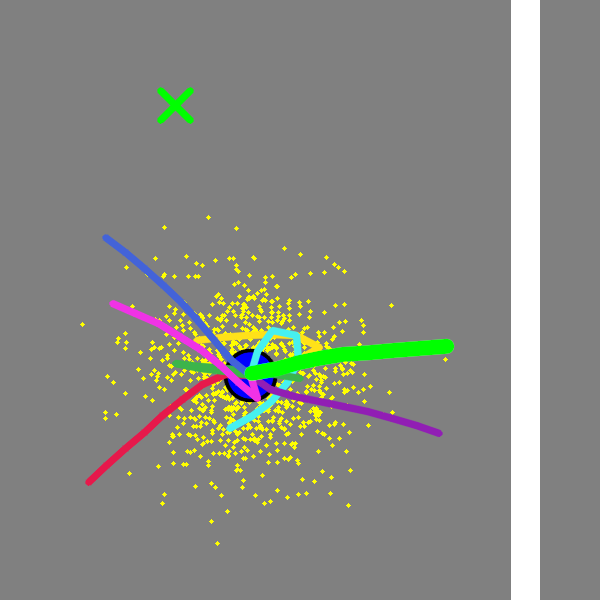}}&
        \Fbox{\includegraphics[width=\linewidth]{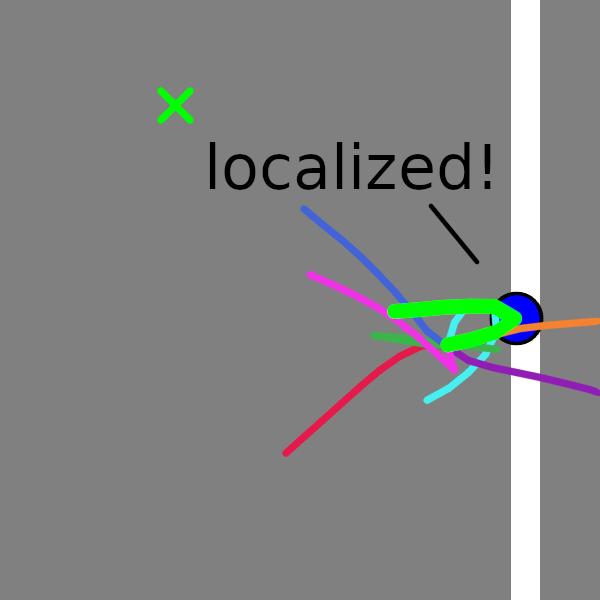}}&
        \Fbox{\includegraphics[width=\linewidth]{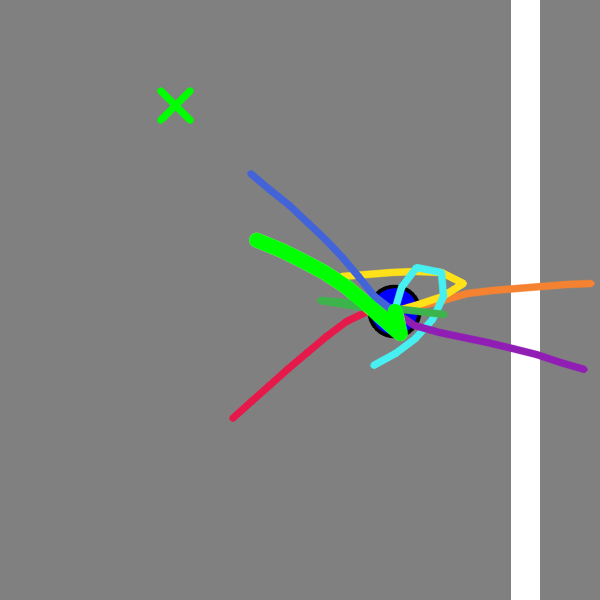}}
        \\
        \begin{subfigure}{80px}\subcaption{\null}\label{lightdarkoverfitd}\end{subfigure}&
        \begin{subfigure}{80px}\subcaption{\null}\label{lightdarkoverfite}\end{subfigure}&
        \begin{subfigure}{80px}\subcaption{\null}\label{lightdarkoverfitf}\end{subfigure}
    \end{tabularx}
    }
    \postsubcapspace
    \caption{Learned macro-actions by the belief-only variant (a-c) and the unconditioned variant (d-f) of \algacro: (a,d) episodes start; (b,e) robot localized; (c,f) goal reaching.}

    \label{lightdarkoverfit}
    \postfigspace
\end{figure}

\subsection{Ablation study -- partially conditioned macro-actions}
We additionally conduct an ablation study (\tabref{tableablation}) to investigate the performance of \algacro when partially conditioning the macro-action generator on only the belief or context. We also compare \algacro with the unconditioned version, which directly learns a fixed macro-action set for all situations using gradient ascent, without the generator.

The results show that \algacro always performs the best when conditioning on \textit{both} the belief and the context (``belief + context''); it typically works the worst when unconditioned on either information (``unconditioned''). 
Specifically, removing belief conditioning (``context only'') harms both Light-Dark and Puck-Push significantly, where the desirable macro-actions depend strongly on the location of the robot/puck and its uncertainty.
Removing context conditioning (``belief only'') causes a significant drop in performance for Light-Dark. This is because the task requires knowing the light position for efficient localization and the goal position to precisely stop at it. 
Without the information, the macro-action set is always scattered even after perfect localization (\figref{lightdarkoverfit}), which leads to less concentrated macro-actions and thus inefficient planning. This phenomenon is observed for both the unconditioned and the belief-only variants. At the goal-reaching phase, the belief-only variant generates particularly unreasonable macro-actions, because the generator overfits to the goal positions in the training data.

\section{Conclusion}
We have presented \algname (\algacro), an algorithm to learn a macro-action generator from data, and to use it to empower long-horizon online POMDP planning. \algacro efficiently learns situation-aware open-loop macro-actions optimized \textit{directly} for planning. By integrating the learned macro-actions with online planning, our planner, \planner, achieves superior performance on various tasks compared to planning with primitive actions and handcrafted macro-actions. We have trained \algacro in simulation and successfully deployed it on a realistic high-fidelity simulator and a real-world setup. More importantly, the core idea of \algacro is generalizable to learning other planner parameters other than macro-action candidates. This would be our future direction.

\section*{Acknowledgements}
This research is supported in part by 
the  A\(^*\)STAR Undergraduate Scholarship (SE/AUS/15/016), the A\(^*\)STAR National Robotics Program (Grant No. 192 25 00054), and the National Research Foundation, Singapore under its AI Singapore Program (AISG Award No. AISG2-RP-2020-016).

\bibliographystyle{plainnat}
\bibliography{root}

\clearpage

\appendix

\subsection{Neural network architectures} \label{appendixnnarchitectures}

\begin{figure}[h]
    \centering
    \includegraphics[width=\linewidth]{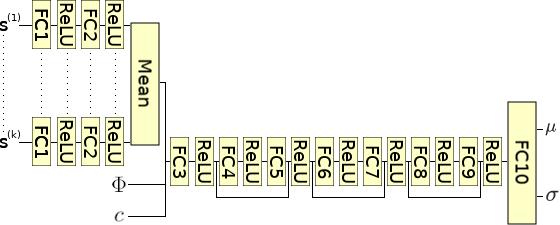}
    \caption{Architecture of \criticnet}
    \label{figurenncriticnet}
\end{figure}

\begin{figure}[h]
    \centering
    \includegraphics[width=\linewidth]{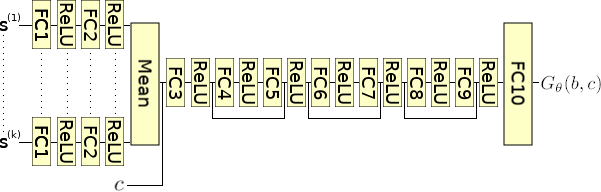}
    \caption{Architecture of \generatornet.}
    \label{figurenngeneratornet}
\end{figure}

\figref{figurenncriticnet} shows the architecture of the \criticnet. It takes real-valued inputs consisting of the current particle-belief \(b\), the context \(\ctx\), and the macro-action set parameters \(\macroactionparams\). Each particle \(s^{(i)}\) in \(b\) is passed through a FC-ReLU stack, with weights shared across all \(s^{(i)}\). This maps the states to a common latent space. The latent particles are then averaged and concatenated with \(\ctx\) and \(\macroactionparams\), and passed through another FC-ReLU stack with residual connections. The final FC layer in \critic outputs a mean and a standard deviation representing a Gaussian distribution over the planner's value estimates.
\figref{figurenngeneratornet} shows the architecture of the \generatornet. It processes the input belief \(b\) using the same parallel FC-ReLU stack as \criticnet. Then, it concatenates the averaged latent particles with the context \(\ctx\), and passes them through another FC-ReLU stack. The final FC output layer produces a real vector corresponding to the parameters of the distribution over \(\macroactionparams\).

\floatstyle{spaceruled}
\restylefloat{algorithm}
\begin{algorithm}[t]
    \caption{\algname (\algacro)}
    \label{ppc}
    \begin{algorithmic}[1]
        \State \(\generatorparams \leftarrow \Call{InitGeneratorNetWeights}{\null}\) \label{ppc:initgeneratornet}
        \State \(\criticparams \leftarrow \Call{InitCriticNetWeights}{\null}\) \label{ppc:initcriticnet}
        \State \(D \leftarrow \Call{InitReplayBuffer}{\null}\) \label{ppc:initbuffer}
        \For{\(i = 1, \dots, W\)} \label{ppc:runworkersstart}
        \State \Call{StartThread}{\Call{Worker}} \label{ppc:runworkers}
        \EndFor
        
        \Function{Worker}{\null}
        \While{\Call{TrainingNotComplete}{\null}} \label{ppc:loopepisode}
        \State \(e, \ctx \leftarrow \Call{InitEnvironment}{\null}\) \label{ppc:epinitstart}
        \State \(b \leftarrow \Call{InitBelief}{\null}\) \label{ppc:epinitend}
        \While {\(\Call{IsNotTerminal}{e}\)} \label{ppc:loopinterleaved}
        \State \(\macroactionparams \sim \generator(b, \ctx)\) \label{ppc:callparametersnet}
        \State \(m, v \sim \Call{Macro-DESPOT}{b, \ctx, \macroactionparams}\) \label{ppc:callplanner}
        \State \(D \leftarrow \Call{Append}{D, (b, \ctx, \macroactionparams, v)}\) \label{ppc:addbuffer}
        \For{\(a\) in \(m = (a_1, \dots)\)} \label{ppc:execstart}
        \State \(e, o \sim \Call{ExecuteAction}{e, a}\) \label{ppc:execend}
        \State \(b \leftarrow \Call{UpdateBelief}{b, a, o}\) \label{ppc:beliefupdate}
        \EndFor
        \State \(b^{(i)}, \ctx^{(i)}, \macroactionparams^{(i)}, v^{(i)} \leftarrow  \Call{Sample}{D, N}\) \label{ppc:samplebuffer}
        \State \(\criticparams \leftarrow \criticparams + \frac{1}{N} \sum_{i = 1}^N \nabla_{\criticparams}J^{(i)}(\criticparams)\) \label{ppc:updatecritic}
        \State \(\generatorparams \leftarrow \generatorparams + \frac{1}{N} \sum_{i = 1}^N \nabla_{\generatorparams}J^{(i)}(\generatorparams)\) \label{ppc:updateparams}
        \State \(\alpha \leftarrow \alpha + \frac{1}{N} \sum_{i = 1}^N \nabla_{\alpha}J^{(i)}(\alpha)\)\label{ppc:updatealpha}
        \EndWhile
        \EndWhile
        \EndFunction
    \end{algorithmic}
\end{algorithm}

\subsection{The training pipeline} \label{appendixtrainingpipeline}

\algref{ppc} illustrates the training pipeline in \algacro. It starts with randomly initialized weights for the generator and critic, \(\generatorparams\) and \(\criticparams\) (Lines \ref{ppc:initgeneratornet}-\ref{ppc:initcriticnet}), together with an empty \textit{replay buffer} (Line \ref{ppc:initbuffer}) to hold experience data. 
It then launches multiple asynchronous workers to collect episodes and perform training (Lines \ref{ppc:runworkersstart}-\ref{ppc:runworkers}). These asynchronous workers share the same neural network weights and the replay buffer, which are access-controlled via mutexes.

A worker resets its environment and its initial belief at the start of each episode (Line \ref{ppc:epinitstart}-\ref{ppc:epinitend}), generating a new context \(\ctx\). Then, it alternates between a planning phase, an execution phase, and a learning phase until the end of the episode. 
In the planning phase, the worker first queries the generator \generator (Line \ref{ppc:callparametersnet}) for the current belief \(b\) and context \(\ctx\) to propose the parameters \macroactionparams of the macro-action set (Line \ref{ppc:callplanner}). \planner then uses \(\macroactionset\) to perform belief tree search; it outputs a macro-action \(m\), together with the quality measure \(v\). The experience \((b, \ctx, \macroactionparams, v)\) is then appended to the replay buffer (Line \ref{ppc:addbuffer}). 
In the execution phase, the worker executes \(m\) as a sequence of primitive actions (Lines \ref{ppc:execstart} to \ref{ppc:execend}), updating the belief (Line \ref{ppc:beliefupdate}) along the way.  
Finally, in the learning phase, a mini-batch of experience \(\{(b^{(i)}, \ctx^{(i)}, \macroactionparams^{(i)}, v^{(i)})\}_i\) is sampled from the replay buffer (Line \ref{ppc:samplebuffer}). It first updates \(\critic\) (Line \ref{ppc:updatecritic}) using Eqn. (\ref{eqncriticobjective}). The updated \(\critic\) is then used to train the generator \(\generator\) (Line \ref{ppc:updateparams}) via Eqn. (\ref{eqnchainrule}). Finally, the entropy regularization weight \(\alpha\) is adjusted (Line \ref{ppc:updatealpha}) via Eqn. (\ref{alphaupdate}).

The three phases repeat until the end of the episode (Line \ref{ppc:loopinterleaved}), after which the next episode starts. The worker repeats the loop (Line \ref{ppc:loopepisode}) until training terminates. In our experiments, we terminate \algacro after reaching a certain number of updates to the neural networks (Lines \ref{ppc:updatecritic}-\ref{ppc:updateparams}).

\begin{table*}[t]
    {\centering
    \caption{Performance (cumulative reward) of MAGIC over different conditionings and macro-action lengths.}
\begin{tabular}{ccccccccccccc}
\toprule
 &
  \multicolumn{4}{c}{Light-Dark} &
  \multicolumn{4}{c}{\begin{tabular}[c]{@{}c@{}}Crowd-Driving\\ (POMDP Simulator)\end{tabular}} &
  \multicolumn{4}{c}{\begin{tabular}[c]{@{}c@{}}Puck-Push\\ (POMDP Simulator)\end{tabular}} \\
  \cmidrule(lr){2-5}\cmidrule(lr){6-9}\cmidrule(lr){10-13}
\begin{tabular}[c]{@{}c@{}}Macro-Action\\ Length\end{tabular} &
  \begin{tabular}[c]{@{}c@{}}Context \\ + Belief\end{tabular} &
  \begin{tabular}[c]{@{}c@{}}Belief\\ Only\end{tabular} &
  \begin{tabular}[c]{@{}c@{}}Context\\ Only\end{tabular} &
  None &
  \begin{tabular}[c]{@{}c@{}}Context \\ + Belief\end{tabular} &
  \begin{tabular}[c]{@{}c@{}}Belief\\ Only\end{tabular} &
  \begin{tabular}[c]{@{}c@{}}Context\\ Only\end{tabular} &
  None &
  \begin{tabular}[c]{@{}c@{}}Context \\ + Belief\end{tabular} &
  \begin{tabular}[c]{@{}c@{}}Belief\\ Only\end{tabular} &
  \begin{tabular}[c]{@{}c@{}}Context\\ Only\end{tabular} &
  None \\ \midrule
2 &
  \begin{tabular}[c]{@{}c@{}}-91.2\\ (1.0)\end{tabular} &
  \begin{tabular}[c]{@{}c@{}}-93.1\\ (1.0)\end{tabular} &
  \begin{tabular}[c]{@{}c@{}}-95.5\\ (1.0)\end{tabular} &
  \begin{tabular}[c]{@{}c@{}}-89.6\\ (1.0)\end{tabular} &
  \begin{tabular}[c]{@{}c@{}}40.4\\ (2.0)\end{tabular} &
  \begin{tabular}[c]{@{}c@{}}30.5\\ (2.0)\end{tabular} &
  \begin{tabular}[c]{@{}c@{}}36.5\\ (2.0)\end{tabular} &
  \begin{tabular}[c]{@{}c@{}}-0.2\\ (2.0)\end{tabular} &
  \begin{tabular}[c]{@{}c@{}}79.7\\ (1.0)\end{tabular} &
  \begin{tabular}[c]{@{}c@{}}73.0\\ (1.0)\end{tabular} &
  \begin{tabular}[c]{@{}c@{}}45.6\\ (1.0)\end{tabular} &
  \begin{tabular}[c]{@{}c@{}}53.0\\ (1.0)\end{tabular} \\
3 &
  \begin{tabular}[c]{@{}c@{}}-92.1\\ (1.0)\end{tabular} &
  \begin{tabular}[c]{@{}c@{}}-92.3\\ (1.0)\end{tabular} &
  \begin{tabular}[c]{@{}c@{}}-88.7\\ (1.0)\end{tabular} &
  \begin{tabular}[c]{@{}c@{}}-100.4\\ (1.0)\end{tabular} &
  \textbf{\begin{tabular}[c]{@{}c@{}}58.6\\ (2.0)\end{tabular}} &
  \textbf{\begin{tabular}[c]{@{}c@{}}53.7\\ (2.0)\end{tabular}} &
  \textbf{\begin{tabular}[c]{@{}c@{}}56.6\\ (2.0)\end{tabular}} &
  \begin{tabular}[c]{@{}c@{}}7.7\\ (2.0)\end{tabular} &
  \begin{tabular}[c]{@{}c@{}}74.2\\ (1.0)\end{tabular} &
  \textbf{\begin{tabular}[c]{@{}c@{}}82.5\\ (1.0)\end{tabular}} &
  \begin{tabular}[c]{@{}c@{}}34.8\\ (1.0)\end{tabular} &
  \begin{tabular}[c]{@{}c@{}}58.8\\ (1.0)\end{tabular} \\
4 &
  \begin{tabular}[c]{@{}c@{}}-75.9\\ (1.0)\end{tabular} &
  \begin{tabular}[c]{@{}c@{}}-82.6\\ (1.0)\end{tabular} &
  \begin{tabular}[c]{@{}c@{}}-64.2\\ (1.0)\end{tabular} &
  \begin{tabular}[c]{@{}c@{}}-30.4\\ (1.0)\end{tabular} &
  \begin{tabular}[c]{@{}c@{}}33.6\\ (2.0)\end{tabular} &
  \begin{tabular}[c]{@{}c@{}}30.1\\ (2.0)\end{tabular} &
  \begin{tabular}[c]{@{}c@{}}33.8\\ (2.0)\end{tabular} &
  \begin{tabular}[c]{@{}c@{}}-12.7\\ (2.0)\end{tabular} &
  \begin{tabular}[c]{@{}c@{}}81.1\\ (1.0)\end{tabular} &
  \begin{tabular}[c]{@{}c@{}}55.3\\ (1.0)\end{tabular} &
  \begin{tabular}[c]{@{}c@{}}54.1\\ (1.0)\end{tabular} &
  \textbf{\begin{tabular}[c]{@{}c@{}}64.3\\ (1.0)\end{tabular}} \\
5 &
  \begin{tabular}[c]{@{}c@{}}-63.9\\ (1.0)\end{tabular} &
  \begin{tabular}[c]{@{}c@{}}-69.4\\ (1.0)\end{tabular} &
  \begin{tabular}[c]{@{}c@{}}-48.2\\ (1.0)\end{tabular} &
  \begin{tabular}[c]{@{}c@{}}-3.5\\ (1.0)\end{tabular} &
  \begin{tabular}[c]{@{}c@{}}43.6\\ (2.0)\end{tabular} &
  \begin{tabular}[c]{@{}c@{}}39.8\\ (2.0)\end{tabular} &
  \begin{tabular}[c]{@{}c@{}}42.6\\ (2.0)\end{tabular} &
  \textbf{\begin{tabular}[c]{@{}c@{}}15.9\\ (2.0)\end{tabular}} &
  \textbf{\begin{tabular}[c]{@{}c@{}}87.9\\ (1.0)\end{tabular}} &
  \begin{tabular}[c]{@{}c@{}}62.1\\ (1.0)\end{tabular} &
  \textbf{\begin{tabular}[c]{@{}c@{}}66.2\\ (1.0)\end{tabular}} &
  \begin{tabular}[c]{@{}c@{}}24.6\\ (1.0)\end{tabular} \\
6 &
  \begin{tabular}[c]{@{}c@{}}-24.5\\ (1.0)\end{tabular} &
  \begin{tabular}[c]{@{}c@{}}-68.8\\ (1.0)\end{tabular} &
  \begin{tabular}[c]{@{}c@{}}-22.2\\ (1.0)\end{tabular} &
  \begin{tabular}[c]{@{}c@{}}22.2\\ (1.0)\end{tabular} &
  \begin{tabular}[c]{@{}c@{}}23.7\\ (2.0)\end{tabular} &
  \begin{tabular}[c]{@{}c@{}}24.0\\ (2.0)\end{tabular} &
  \begin{tabular}[c]{@{}c@{}}20.5\\ (2.0)\end{tabular} &
  \begin{tabular}[c]{@{}c@{}}11.3\\ (2.0)\end{tabular} &
  \begin{tabular}[c]{@{}c@{}}65.9\\ (1.0)\end{tabular} &
  \begin{tabular}[c]{@{}c@{}}13.7\\ (1.0)\end{tabular} &
  \begin{tabular}[c]{@{}c@{}}49.0\\ (1.0)\end{tabular} &
  \begin{tabular}[c]{@{}c@{}}46.9\\ (1.0)\end{tabular} \\
7 &
  \begin{tabular}[c]{@{}c@{}}22.1\\ (1.0)\end{tabular} &
  \begin{tabular}[c]{@{}c@{}}-52.9\\ (1.0)\end{tabular} &
  \textbf{\begin{tabular}[c]{@{}c@{}}35.1\\ (1.0)\end{tabular}} &
  \begin{tabular}[c]{@{}c@{}}42.7\\ (1.0)\end{tabular} &
  \begin{tabular}[c]{@{}c@{}}24.0\\ (2.0)\end{tabular} &
  \begin{tabular}[c]{@{}c@{}}18.8\\ (2.0)\end{tabular} &
  \begin{tabular}[c]{@{}c@{}}20.1\\ (2.0)\end{tabular} &
  \begin{tabular}[c]{@{}c@{}}11.0\\ (2.0)\end{tabular} &
  \begin{tabular}[c]{@{}c@{}}60.2\\ (1.0)\end{tabular} &
  \begin{tabular}[c]{@{}c@{}}-13.1\\ (1.0)\end{tabular} &
  \begin{tabular}[c]{@{}c@{}}16.9\\ (1.0)\end{tabular} &
  \begin{tabular}[c]{@{}c@{}}15.8\\ (1.0)\end{tabular} \\
8 &
  \textbf{\begin{tabular}[c]{@{}c@{}}54.1\\ (1.0)\end{tabular}} &
  \textbf{\begin{tabular}[c]{@{}c@{}}-51.2\\ (1.0)\end{tabular}} &
  \begin{tabular}[c]{@{}c@{}}-15.6\\ (1.0)\end{tabular} &
  \textbf{\begin{tabular}[c]{@{}c@{}}43.9\\ (1.0)\end{tabular}} &
  \begin{tabular}[c]{@{}c@{}}2.1\\ (2.0)\end{tabular} &
  \begin{tabular}[c]{@{}c@{}}1.9\\ (2.0)\end{tabular} &
  \begin{tabular}[c]{@{}c@{}}-5.1\\ (2.0)\end{tabular} &
  \begin{tabular}[c]{@{}c@{}}-3.3\\ (2.0)\end{tabular} &
  \begin{tabular}[c]{@{}c@{}}34.0\\ (1.0)\end{tabular} &
  \begin{tabular}[c]{@{}c@{}}27.8\\ (1.0)\end{tabular} &
  \begin{tabular}[c]{@{}c@{}}-1.5\\ (1.0)\end{tabular} &
  \begin{tabular}[c]{@{}c@{}}-13.8\\ (1.0)\end{tabular} \\ \bottomrule
\end{tabular}
    
    \label{tableablationfull}
    }
\end{table*}

\subsection{Macro-action parameterizations} \label{appendixmacroparameterization}
\begin{figure}[h]
    \centering
    \begin{tabular}{cc}
        \includegraphics[width=0.4\linewidth]{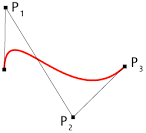} &      \includegraphics[width=0.4\linewidth]{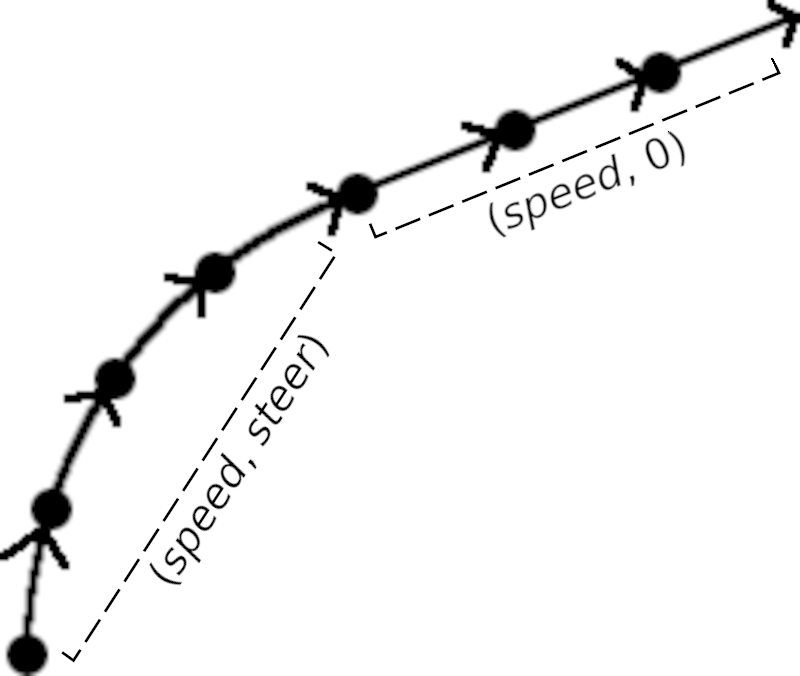}\\
        \begin{subfigure}{40px}\subcaption{\null}\label{beziercurve}\end{subfigure}&
        \begin{subfigure}{40px}\subcaption{\null}\label{turnandgocurve}\end{subfigure}
    \end{tabular}
    \caption{Macro-action representations: (a) 2D Bezier curves for Light-Dark and Puck-Push, and (b) turn-and-go curves for Crowd-Driving.}
\end{figure}

\figref{beziercurve} shows an example of the macro-action parameterization used for 2D navigation tasks, \ie, Light-Dark and Puck-Push. A macro-action is represented as a 2D Bezier curve \cite{mortenson1999beziercurve}, defined using three control points \(P_1, P_2, P_3\). The corresponding trajectory is a curve bounded within the convex hull of the control points. 
\figref{turnandgocurve} illustrates the turn-and-go curve used for Crowd-Driving, characterized by two real parameters, \(speed\) and \(steer\). Here, $speed$ denotes the command speed across the entire trajectory, increasing which results in longer trajectories; $steer$ specifies the steering angle used in the first half of the curve, increasing which produces sharper turns.


\subsection{Details of Evaluation Tasks} \label{appendixtaskdetails}
\subsubsection{Light-Dark}
A mobile robot operates in a 2D world. At every step, the robot can either move along an arbitrary direction for a fixed distance, or \textit{STOP} the robot, hopefully at the goal. A vertical light area exists in the world, within which the robot receives accurate readings of its position, and outside which the robot receives completely no observation of its position. The robot receives a reward of \(-0.1\) at every step, \(+100\) for stopping correctly at the goal, and \(-100\) for stopping incorrectly. Otherwise, \(STOP\) is automatically executed when exhausting the steps allowed. 
Each episode allows a maximum of \(60\) steps. Planners use a maximum search depth of \(60\), with \(\SI{0.1}{\second}\) of maximum planning time.

\subsubsection{Crowd-Driving}
A robot vehicle drives along a target path at a busy intersection through crowded traffic. The robot can observe the positions, orientations, velocities of all vehicles including the ego-agent's, but not their intentions. The robot is controlled by speed and steering commands, with the maximum speed, steering, and acceleration constrained to be \(\SI{6.0}{\meter/\second}\), \(15^\circ\), and \(\SI{3.0}{\meter/\second\squared}\). Exo-vehicles in the crowd are modeled using GAMMA \cite{luo2019gamma}, a recent traffic motion model based on Reciprocal Velocity Obstacles (RVO). Exo-agents are also controlled by speed and steering, with the maximum speed, steering, and acceleration constrained to be \(\SI{4.0}{\meter/\second}\), \(15^\circ\), and \(\SI{2.0}{\meter/\second\squared}\). At each step, the robot receives a reward equal to the distance traveled along the target path, and receives a penalty of \(-3.61\) if the robot's speed falls below half its maximum speed.
The episode terminates if a collision occurs, and the robot receives a collision penalty of \(-100\). 
A maximum of \(150\) steps is allowed and a planning depth of \(40\) is used. We execute each macro-action while concurrently planning for the next step. 
The amortized planning rate is \(\SI{5}{\hertz}\).

\subsubsection{Puck-Push}
A mobile robot pushes a puck to a goal in a 2D workspace. 
When being pushed, the puck passively slides along the robot's circumference. The sliding motion is modeled by \(\theta' = \theta e^{\mu d}\), where \(\theta\) is the direction of the puck w.r.t. to the robot's ego-frame, \(d\) is the distance the robot moved during contact, and \(\mu\) is a sliding rate coefficient determined by physical properties of the puck. Gaussian noises are added to the motion of both the robot and the puck to simulate uncertain transitions.
Observations come from a bird's-eye view camera. When the puck passes the yellow occluding regions, the camera provides no observation. Outside the regions, the perception pipeline also fails occasionally, which is modeled by omitting observations with probability \(0.1\). The robot receives a reward of \(-0.1\) per step, \(+100\) for successful delivery, and \(-100\) for exhausting the allowed steps or colliding with the boundary.  
Each episode allows a maximum of 100 steps. Planners use a maximum search depth of 100, and are given \(\SI{0.1}{\second}\) of planning time.

\subsection{Detailed results over conditioning and macro-action length} \label{appendixdetailedresults}
\tabref{tableablationfull} shows the performance of \algacro using different macro-action lengths. Performance generally improves with increasing macro-action length up to a certain length, after which performance degrades again. This is because short macro-actions do not cover enough depth for long-horizon planning, while very long macro-actions are hard to learn using gradient ascent.

\vfill\eject

\end{document}